%% file: main.tex
\documentclass[11pt]{article}
\usepackage{subfigure}
\input{Preamb.tex}

\usepackage[numbers,merge,sort&compress]{natbib}
\begin{document}

\title{
 \bf 
 \Large Model-free Online Learning for the Kalman Filter: Forgetting Factor and Logarithmic Regret
\thanks{This work is supported by NSF ECCS 2154650, NSF CMMI 2320697, and NSF CAREER 2340713. Emails: jiq012@ucsd.edu; zhengy@ucsd.edu. } 
}
\author[1]{Jiachen Qian}
\author[1]{Yang Zheng}
\affil[1]{\small Department of Electrical and Computer Engineering, University of California San Diego}
\date{\small \today \vspace{-5ex}} 

\maketitle

\begin{abstract}
    We consider the problem of online prediction for an unknown, non-explosive linear stochastic system.
With a known system model, the optimal predictor is the celebrated Kalman filter. 
In the case of unknown systems, existing approaches based on recursive least squares and its variants may suffer from degraded performance due to the highly imbalanced nature of the regression model. This imbalance can easily lead to overfitting and thus degrade prediction accuracy.
We tackle this problem by injecting an inductive bias into the regression model via {exponential forgetting}. While exponential forgetting is a common wisdom in online learning, it is typically used for
re-weighting data.  In contrast, our approach focuses on balancing the regression model. This achieves a better trade-off between {regression} and {regularization errors}, and simultaneously reduces the {accumulation error}. With new proof techniques, we also provide a sharper logarithmic regret bound of $\mathcal{O}(\log^3 N)$, where $N$ is the number of observations.   
\end{abstract}

\input{Section-I-Intro-v1}
\input{Section-II-v1.tex}

\input{Section-III-v1}
\input{Section-V-v0}

\section{Conclusion} \label{section:conclusion}
In this paper, we addressed the problem of online prediction for an unknown, non-explosive linear stochastic system. 
To counteract the imbalanced regression model, which degrades the performance of existing prediction algorithms, we introduced an exponential forgetting strategy in our \texttt{OPF} algorithm that injects an inductive bias. 
This approach leads to a more balanced regression model and helps mitigate overfitting. 

Using refined proof techniques, we developed a structured regret analysis framework, particularly for marginally stable systems,  
and we established a sharper regret bound as $O(\log^3 N)$. 
Our findings highlight the importance of balancing the regression model rather than just re-weighting past data. Future directions include exploring adaptive forgetting strategies and extending the framework to nonlinear systems, such as special classes of structured nonlinear models \cite{shang2024willems}.

{ 
\bibliographystyle{unsrt}
\bibliography{reference.bib}
}

\small 
\newpage
\appendix
\numberwithin{equation}{section}

\noindent \textbf{\Large Appendix}
\vspace{3mm}

In this appendix, we provide extra discussions, proof details, and further numerical results. In particular, we organize this appendix as follows:
\begin{itemize}
    \item \Cref{secLinear} provides some basic statistical properties of linear stochastic systems, including high probability upper and lower bounds for system states, prediction errors, Gram matrices, etc.
    
    \item \Cref{secBackground} provides some basic background theory, such as the Kalman filter and Hanson-Wright inequality, which are essential for the proof of logarithm regret.
    \item \Cref{secMainProof} provides the detailed proof for \Cref{thm1}.

    \item \cref{secDiscussion} provides a detailed discussion on the effect of the forgetting factor $\gamma$ to improve the prediction performance and the robustness to guarantee the regret $\mathcal{R}_N$ to be sub-linear.
    
    \item \Cref{secAdditionalExperiments} provides supplementary numerical experiments including further discussion on the prediction algorithm, theoretical verifications and comparison with related algorithms in \cite{rashidinejad2020slip}. 
    \item \Cref{secFurther} discusses further extensions of the theoretical results, including the case with control input. 
\end{itemize}

\vspace{5mm}

\noindent \textbf{Notation.} We provide the frequently-used notations of this paper in Table \ref{tableNotation}. In the appendix, to provide a more concise proof, we will use one term $M=\operatorname{poly}(\left\|A\right\|_2,\left\|C\right\|_2,\left\|Q\right\|_2,\left\|R\right\|_2)$ to represent an uniform upper bound for all system related parameters.  
\begin{table}[h]
\caption{Frequently-used Notations}
\label{tableNotation}
\vspace{-5mm}
\begin{center}
\begin{small}
\begin{sc}
\renewcommand{\arraystretch}{1.3} 
\begin{tabular}{lcr}
\toprule
\textnormal{Notation} & \textnormal{Meaning} &\\
\midrule
$Z_{k,p}=\begin{bmatrix}
    y_{k-p}^{\tr}&\cdots &y_{k-1}^{\tr}
\end{bmatrix}^{\tr}$&\textnormal{Sample at time step} $k$\\

$\bar{Z}_{k,p}=\begin{bmatrix}
    Z_{p,p} & \cdots & Z_{k,p}
\end{bmatrix}$ & \textnormal{Collected samples up to time step} $k$\\

$Y_{k,p}=\begin{bmatrix}
    y_{p} & \cdots & y_{k}
\end{bmatrix}$ 
& \textnormal{Collected observations up to time step}  $k$ \\

$E_{k,p}=\begin{bmatrix}
    e_{p} & \cdots & e_{k}
\end{bmatrix}$     & \textnormal{Collected innovation up to time step} $k$ \\

$b_{k,p}=C\left(A-LC\right)^p\hat{x}_{k-p}$     & \textnormal{Truncation bias at time step} $k$ \\

$B_{k,p}=\begin{bmatrix}
    b_{p,p} & \cdots & b_{k,p}
\end{bmatrix}$     & \textnormal{Collected truncation bias up to time step} $k$ \\

$D_{p}=\operatorname{diag}\begin{bmatrix}
    \gamma^{p-1} & \cdots & 1
\end{bmatrix}\otimes I_m$     & \textnormal{Scaling matrix induced by our forgetting factor} \\

$\tilde{Z}_{k,p}=D_pZ_{k,p}$     & \textnormal{Scaled sample at time step} $k$ \\

$\tilde{V}_{k,p}=\lambda I+D_p\bar{Z}_{k, p}\bar{Z}_{k, p}^{\tr}D_p$
   & \textnormal{Gram matrix at time step} $k$        \\

$\bar{V}_{k,p}=\lambda D_p^{-2}+\bar{Z}_{k, p}\bar{Z}_{k, p}^{\tr}$      & \textnormal{Equivalent form of Gram matrix for proof}\\

$\bar{R}\triangleq C\bar{P}C^{\tr}+R$& \textnormal{Covariance of steady-state innovation} \\

$\Gamma_k\triangleq \mathbb{E}\left\{\hat{x}_k\hat{x}_k^{\tr}\right\}$& \textnormal{Second moment of estimation} $\hat{x}_k$\\

$\Gamma_{k,p}^{Z}\triangleq
\mathbb{E}\left\{Z_{k,p}Z_{k,p}^{\tr}\right\}$& \textnormal{Second moment of sample} $Z_{k,p}$\\
$\mathcal{E}_x,\mathcal{E}_Z,\mathcal{E}_e,\mathcal{E}_{PE}$& \textnormal{High Probability event for the formulation of regret bound}\\
$M=\operatorname{poly}(\left\|A\right\|_2,\left\|C\right\|_2,\left\|Q\right\|_2,\left\|R\right\|_2)$& 
\textnormal{Uniform upper bound only related to system parameters}\\
$\sigma_R, \bar{\sigma}_{\bar{R}}$& \textnormal{Minimal singular value of $R$ and maximal singular value of $\bar{R}$}\\
\bottomrule
\end{tabular}
\end{sc}
\end{small}
\end{center}
\vskip -0.1in
\end{table}

Moreover, we denote $\succ$ as comparison in the positive semi-definite cone. We have that $\left\|\cdot\right\|_2$ denotes the 2-norm, $\left\|\cdot\right\|_F$ denotes the F-norm and $\left\|\cdot\right\|_1$ denotes the 1-norm. $\left\|X\right\|_{\psi_2}$ denotes the $\psi_2$ norm of sub-Gaussian random variable \citep[Definition 2.5.6]{vershynin2018high}. $\rho(A)$ denotes the spectral radius of matrix $A$. $\operatorname{poly}(x)$ denotes a polynomial of elements $x$. $O(f(x))$ indicates the function is with the same order as $f(x)$ and $\tilde{O}(f(x))$ indicates the function is with the same order as $f(x)\log^n(f(x))$. 

\section{Linear Stochastic System} \label{secLinear}

In this section, we introduce some basic properties of the linear stochastic system. Consider the following linear system model
\begin{equation}\label{linearsystem}
\begin{aligned}
x_{k+1}&=Ax_k + \omega_k,\\
y_{k}&=Cx_k + v_{k},\quad k = 0,1,2,\ldots
\end{aligned}
\end{equation}
where $x_k\in\mathbb{R}^n$ is the state vector of the system and $y_{k}\in\mathbb{R}^{m}$ is the measurement vector,  $\omega_k\in\mathbb{R}^n$ is the process noise with covariance matrix $Q\in\mathbb{R}^{n\times n}$ and $v_{k}\in\mathbb{R}^{m}$ is the observation noise with covariance matrix  $R\in\mathbb{R}^{m \times m}$. The sequences $\left\{\omega_k\right\}^{\infty}_{k=0}$ and $\left\{v_{k}\right\}^{\infty}_{k=0}$ are {mutually uncorrelated white Gaussian noise}. $Q$ and $R$ are positive definite. Besides, $A$ is the state-transition matrix and $C$ is the observation matrix. 
For the theoretical analysis of the regret, we present the following basic properties of the linear stochastic system. 

\subsection{Basic statistical properties of random variables}

In this subsection, we introduce some basic statistical properties about the state predictions $\hat{x}_k$, samples $Z_{k,p}$, and innovations $e_k$, which are essential for the following proof.
We first define the following notations,
$$
\bar{R}=CPC^{\tr}+R, \quad\Gamma_k\triangleq\mathbb{E}\left\{\hat{x}_k\hat{x}_k^{\tr}\right\},\quad \Gamma^{Z}_{k,p}\triangleq\mathbb{E}\left\{Z_{k,p}Z_{k,p}^{\tr}\right\},
$$
and the following events
$$
\begin{aligned}
    \mathcal{E}_{x} \triangleq&\left\{\left\|\Gamma_{k}^{-1 / 2} \hat{x}_{k}\right\|_{2} \leq \sqrt{n}+\sqrt{2 \log \frac{4k}{\delta}}, \;\;\forall k\ge T_{\text{init}}\right\},\\
    \mathcal{E}_{Z} \triangleq&\left\{\left\|\left(\Gamma_{k,p}^Z\right)^{-1 / 2} Z_{k,p}\right\|_{2} \leq \sqrt{mp}+\sqrt{2 \log \frac{4 k}{\delta}}, \;\; \forall k\ge T_{\text{init}}\right\},\\
    \mathcal{E}_{e}\triangleq&\left\{\left\|\bar{R}^{-1 / 2} e_{k}\right\|_{2} \leq \sqrt{m}+\sqrt{2 \log \frac{4 k}{\delta}}, \;\; \forall k\ge T_{\text{init}}\right\}.
\end{aligned}
$$
We have the following Lemma.
\begin{lemma}\label{lmDeviation}
    For any given $\delta\in (0,1)$, each of event $\mathcal{E}_{x}$,$\;\mathcal{E}_{e}$ and $\mathcal{E}_{Z}$ holds with probability at least $1-\frac{\pi^2 \delta}{6}$.
\end{lemma}
\begin{proof}
     From Lemma 1 in \cite{laurent2000adaptive}, 
     for any Gaussian random vector $X_k\sim \mathcal{N}(0,I_n).$ we have
     $$\mathbb{P}\left(\left\|X_k\right\|_2^2\ge n+2\sqrt{n}t+2t^2\right)\leq e^{-t^2}.$$
     Then the following statement also holds
     $$
     \mathbb{P}\left(\left\|X_k\right\|_2^2\ge \left(\sqrt{n}+\sqrt{2}t\right)^2\right)\leq e^{-t^2}.
     $$
     For each time step $k$, take $t=\sqrt{\log\frac{k^2}{\delta}}$, then we have$$\mathbb{P}\left(\left\|X_k\right\|_2\leq \sqrt{n}+\sqrt{4\log\frac{k}{\delta}}\right)\ge 1- \frac{\delta}{k^2}.
     $$

     Take a union bound over all $k$, we have
     $$
     \mathbb{P}\left(\left\|X_k\right\|_2\leq \sqrt{n}+\sqrt{4\log\frac{k}{\delta}}, \forall k\in\mathbb{N}\right)\ge 1- \sum_{k=1}^{\infty}\frac{\delta}{k^2}=1-\frac{\pi^2\delta}{6},
     $$
     where the inequality utilizes the facts 
     $$\mathbb{P}(\mathcal{E}_1 \cap \mathcal{E}_2)=\mathbb{P}(\mathcal{E}_1)+\mathbb{P}(\mathcal{E}_2)-\mathbb{P}(\mathcal{E}_1 \cup \mathcal{E}_2)$$ and $\mathbb{P}(\mathcal{E}_1 \cup \mathcal{E}_2)\leq 1$.
     Note that $\Gamma_k^{-1/2}\hat{x}_k\sim \mathcal{N}(0,I_n)$, hence with the above argument, we have
     $\mathbb{P}(\mathcal{E}_x)\leq 1-\frac{\pi^2\delta}{6}$. 
     
     The other two arguments can be similarly proved.
\end{proof}
The above lemma illustrates that due to the zero-mean Gaussianity of the random variable $\hat{x}_k,Z_{k,p}$ and $e_k$, we can find an increasing upper bound with the same order as $O(\log k)$ for the deviation between the random variables and their corresponding expectation with high probability $1-\delta$. This lemma will be repeatedly used in the analysis of the regret bound.

\subsection{Upper bound of $\bar{Z}_{k,p}\bar{Z}_{k,p}^{\tr}$ with high probability}\label{upperbound}
The following lemma provides an upper bound for $\bar{Z}_{k,p}\bar{Z}_{k,p}^{\tr},\forall k\ge T_{\text{init}}$.
\begin{lemma}
    For a given probability $\delta$, we have $\mathbb{P}\left\{\bar{Z}_{k,p}\bar{Z}_{k,p}^{\tr}\leq \operatorname{poly}\left(M,m,\beta,\log\frac{1}{\delta}\right)k^{2\kappa+1} I_{mp}\right\}\ge 1-\delta$, where $\operatorname{poly}(\cdot)$ means polynomial operator.
\end{lemma}
\begin{proof}
   For the expectation of Gram matrix $Z_{k,p}Z_{k,p}^{\tr}$, we have
$$
\begin{aligned}
    \left\|\Gamma_{k,p}^Z\right\|_2^2\leq\mathbb{E}\left\{Z_{k,p}^{\tr}Z_{k,p}\right\}
    =&\mathrm{tr}\left(\sum_{i=k-p}^{k-1}\mathbb{E}\left\{y_{i}y_{i}^{\tr}\right\}\right)\\
    =&p\mathrm{tr}(R)+\text{tr}\left(C^{\tr}C\sum_{i=k-p}^{k-1}\sum_{l=0}^{i-1}A^lQ(A^l)^{\tr}\right)\\
    \leq& mp\left(\left\|R\right\|_2+\left\|C\right\|_2^2\left\|Q\right\|_2\sum_{l=0}^{k-1}\left\|A^l\right\|_2^2\right).
\end{aligned}
$$
Note that exists $M_1$, such that $\left\|A^l\right\|_2\leq M_1l^{\kappa-1}\rho(A)^{l}$, where $\kappa$ is the order of the largest Jordan block of  matrix $A$. Then we have 
$$
\mathbb{E}\left\{Z_{k,p}^TZ_{k,p}\right\}\leq mpM k^{2\kappa-1},\quad \forall k\in\mathbb{N}.
$$
Conditioned on event $\mathcal{E}_Z$, we have
$$
Z_{k,p}Z_{k,p}^{{\tr}}\leq mpM\left(2mp+4\log\frac{4k}{\delta}\right)k^{2\kappa-1} I.
$$
Together with the selection rule of $p\leq \beta \log k$, we can obtain 
$$
\bar{Z}_{k,p}\bar{Z}_{k,p}^{\tr}=\sum_{l=p}^{k}Z_{l,p}Z_{l,p}^{\tr}\leq \operatorname{poly}\left(M,m,\beta,\log\frac{1}{\delta}\right)k^{2\kappa
+1}I
$$
with probability $1-\frac{\pi^2 \delta}{6}$. We now complete the proof. 
\end{proof}

\subsection{Persistent excitation for $\bar{Z}_{k,p}\bar{Z}_{k,p}^{\tr}$ with high probability}
In this subsection, we will provide a lower bound for $\bar{Z}_{k,p}\bar{Z}_{k,p}^{\tr}$ with high probability, which is also known as the condition of persistent excitation. The property of persistent excitation is of vital importance to guarantee the performance of online prediction \Cref{algPrediction}.
\begin{lemma}\label{persistent}
For a given system model \eqref{linearsystem}, then for any fixed failure probability $\delta$ and horizon parameter $\beta$, there exists a number $N_0=\operatorname{poly}\left(\beta,\log\frac{1}{\delta}\right)$, such that for any $k>N_0$ and $p\leq\beta \operatorname{log} k$, there is $\bar{Z}_{k,p}\bar{Z}_{k,p}^{\tr}\ge \frac{\sigma_R}{4}kI$ for all $k$ with probability $1-\delta$, where $\sigma_R$ is the smallest eigenvalue of matrix $R$.
\end{lemma}
\begin{proof}
    From Theorem I.1 in \cite{tsiamis9894660}, we have that for any given $p,k,\delta_1$, if 
    $$
    k\ge \max\left\{k_{1}(p, \delta_1),k_{2}(k, p, \delta_1)\right\}
    $$
    where
    $$
    \begin{aligned} 
k_{1}(p, \delta_1) \triangleq&\; C_1pm \log (p m / \delta_1),\\ 
k_{2}(k, p, \delta_1) \triangleq& \frac{C_2 p n}{\min \left\{4, \sigma_{R}\right\}} \log \left(\frac{5 p}{\delta_1} \frac{n\left\|\mathcal{O}_{p}\right\|_{2}^{2}\left\|\Gamma_{k-p}\right\|_{2}+\delta_1}{\delta_1}\right),
\end{aligned}
    $$
    and $C_1,C_2$ are constants not related to the system parameter and time step $k$.
Then with probability at least $1-3\delta_1$, there is 
$$
\bar{Z}_{k,p}\bar{Z}_{k,p}^{\tr}\ge \frac{\sigma_{\bar{R}}}{4}kI.
$$
From Lemma I.1 in \cite{tsiamis9894660},
there is 
$
\left\|\mathcal{O}_{p}\right\|_{2}^{2}\leq M_1^3 O(p^\kappa), 
$
and
$
\left\|\Gamma_{t}\right\|_{2}= M_1^7O(t^{2\kappa-1})
$, where $M_1$ is only related to $\left\|A\right\|_2$. Hence we have
$$
\left\|\mathcal{O}_{p}\right\|_{2}^{2}\leq M p^\kappa,\quad \left\|\Gamma_{t}\right\|_{2}\leq M t^{2\kappa-1}.
$$
Replace the above $\delta_1$ with $\frac{\delta_1}{k^2}$ for each time step $k$, then we have for each $k$, if 
$$
k\ge \max\left\{k_{1}\left(p, \frac{\delta_1}{k^2}\right),k_{2}\left(k, p, \frac{\delta_1}{k^2}\right)\right\},
$$
then with probability $1-\frac{3\delta_1}{k^2}$, the persistent excitation condition holds. 
For the term $k_{1}\left(p, \frac{\delta_1}{k^2}\right)$,
we have
$$
k_{1}\left(p, \frac{\delta_1}{k^2}\right)\leq C_1\beta\log(k)\log\left(\frac{\beta m}{\delta_1}k^2\log(k)\right),
$$
where the parameter $C_1,\beta,m,\delta_1$ are unrelated with time step $k$. It is also easy to verify that
$$
C_1\beta\log(k)\log\left(\frac{\beta m}{\delta_1}k^2\log(k)\right)\leq \left(C_1\beta\log\left(\frac{\beta m}{\delta_1}\right)+3C_1\beta\right)\log^2(k),
$$
which indicates that there exists a number $N_1=\operatorname{poly}(C_1,\beta,m,\log(1/\delta_1))$, such that for all $k\ge N_1$, there is
$$
k\ge C_1\beta\log(k)\log\left(\frac{\beta m}{\delta_1}k^2\log(k)\right)\ge k_{1}\left(p, \frac{\delta_1}{k^2}\right).
$$

For the term $k_{2}\left(k, p, \frac{\delta_1}{k^2}\right)$, there is 
$$
\begin{aligned}
    k_{2}\left(k, p, \frac{\delta_1}{k^2}\right) \leq &\frac{C_2 n\beta \log(k)}{\min \left\{4, \sigma_{R}\right\}} \left(\log \left(\frac{5 \beta k^2\log(k)}{\delta_1}\right)+\log\left( \frac{Mnk^2p^\kappa\left\|\Gamma_{k-p}\right\|_{2}}{\delta_1}+1\right)\right)\\
    \leq&\frac{C_2 n\beta \log(k)}{\min \left\{4, \sigma_{R}\right\}} \left(\log \left(\frac{5 \beta k^2\log(k)}{\delta_1}\right)+\log\left( \frac{M^2\beta^\kappa nk^{2\kappa+1}\log^\kappa(k)}{\delta_1}+1\right)\right).
\end{aligned}
$$
With the same argument as the analysis for $k_1$, there exists a number 
$N_2=\operatorname{poly}(C_2,\beta,1/\sigma_R,n,M_1,\allowbreak \log(1/\delta_1),\kappa)$, such that for all $k\ge N_2$, there is 
$$
k\ge \frac{C_2 n\beta \log(k)}{\min \left\{4, \sigma_{R}\right\}} \left(\log \left(\frac{5 \beta k^2\log(k)}{\delta_1}\right)+\log\left( \frac{M_1^2\beta^\kappa nk^{2\kappa+1}\log^\kappa(k)}{\delta_1}+1\right)\right).$$ 
Note that $n,m,\kappa,C_1,C_2,M$ are system related parameters. With the previous discussion, we have for any $k\ge \max\left\{N_1,N_2\right\}$ with probability at least 
$1-3\sum_{k=1}^{\infty}{\frac{1}{k^2}}\delta_1$, there is 
$$
\bar{Z}_{k,p}\bar{Z}_{k,p}^{\tr}\ge \frac{\sigma_{R}}{4}kI.
$$
Then we only need to choose $\delta_1 = \frac{2}{\pi^2}\delta$ and $N_0=\max\left\{N_1,N_2\right\}.$
\end{proof}

This lemma illustrates that with the given parameters $A,C,Q,R,\beta,\delta$, we can find a sufficiently large number $N_0$, such that the persistent excitation condition holds for all $k\ge N_0$. Note that $N_0=\operatorname{poly}\left(\beta,\log\left(\frac{1}{\delta}\right)\right)$ is polynomial to the above parameter, then we can choose the $T_{\text{init}}$ to be large enough to guarantee the persistent excitation condition of the Gram matrix $\bar{V}_{k,p}$  with high probability. In the following discussion, in order to simplify the analysis, we will directly assume that there is $T_{\text{init}}>N_0$.

\section{Background on Kalman filter and Hanson-Wright inequality}\label{secBackground}

\subsection{The Kalman Filter}

Consider the stochastic system \eqref{linearsystem} and the following assumption
\begin{assumption}\label{aspOb}
    The system pair $(A, C)$ is detectable.
\end{assumption}

Let  $\mathcal{F}_{k} \triangleq \sigma\left(y_{0}, \ldots, y_{k}\right)$  be the filtration generated by the observations  $y_{0}, \ldots, y_{k}$. Given the observations up to time $k$, the optimal prediction  $\hat{y}_{k+1}$ at time  $k+1$ in the minimum mean square error sense is defined as
\begin{equation}\label{MMSEProblem}
    \hat{y}_{k+1} \triangleq \arg \min _{z \in \mathcal{F}_{k}} \mathbb{E}\left[\left\|y_{k+1}-z\right\|_{2}^{2} \mid \mathcal{F}_{k}\right] .
\end{equation}
With the Theory of Kalman Filtering \cite{kalmanfilter, Andersonoptimal}, the optimal predictor  of the linear stochastic system \eqref{linearsystem} takes the form as
\begin{equation}\label{OptimalPDT}
\begin{aligned}
\hat{x}_{k+1} & =A \hat{x}_{k}+L_k\left( y_{k}-\hat{y}_{k}\right) 
,\;\;\hat{x}_{0}=0 \\
\hat{y}_{k} & =C \hat{x}_{k}.
\end{aligned}
\end{equation}
where $L_k=AP_kC^{\tr}\left(CP_kC^{\tr}+R\right)^{-1}$, and $P_k$ satisfies the recursion
\begin{equation}\label{RicRecursion}
    P_{k+1}=AP_k A^{\tr}+Q-AP_kC^{\tr}\left(CP_kC^{\tr}+R\right)^{-1}CP_kA^{\tr}.
\end{equation}

From \citep[Appendix E]{kailath2000linear},
with Assumption \ref{aspOb}, we have the following results
\begin{enumerate}
    \item The recursion \eqref{RicRecursion} will converge to steady-state exponentially fast with the increase of the time step $k$. The steady-state performance of the above recursion satisfies the discrete-time algebraic Riccati equation
\begin{equation}
    P=APA^{\tr}+Q-APC^{\tr}\left(CPC^{\tr}+R\right)^{-1}CPA^{\tr}.
\end{equation}
\item Let $L=APC^{\tr}\left(CPC^{\tr}+R\right)^{-1}$, then there is
$\rho(A-LC)<1$.
\end{enumerate}

Denote $e_k=y_k -\hat{y}_k$ as the prediction error, also called innovation, at time step $k$. Then from \citep[P98]{speyer2008stochastic}, we have the following results
\begin{enumerate}
    \item With the basic assumptions of the stochastic system \eqref{linearsystem}, we have
    $$
    \mathbb{E}\left\{e_le_k^{\tr}\right\}=\bar{R}\delta_{l,k},
    $$
    where $\delta_{l,k}=1$ if $l=k$ and $\delta_{l,k} = 0$ if $l\neq k$.
\end{enumerate}

\subsection{Multi-dimensional self-normalizing vector martingales}
In this subsection, we will provide the basic result about Hanson-Wright inequality to provide a tighter bound for the quadratic form of $\left\|E_{k,p}\bar{Z}_{k,p}\bar{V}_{k,p}^{-\frac{1}{2}}\right\|_2^2$.

\subsubsection{Hanson-Wright inequality}

In this subsection, we first introduce the definition of sub-Gaussian random variables and the corresponding Hanson-Wright inequality for the subsequent analysis on the regression error.
\begin{lemma}\label{HWinequality}
(Hanson-Wright inequality \cite{vershynin2018high}) Let  $X=\left(X_{1}, \ldots, X_{n}\right) \in \mathbb{R}^{n}$  be a random vector with independent, mean zero, sub-Gaussian coordinates. Let  $A$  be an  $n \times n$  matrix. Then, for every  $t \geq 0$, we have
    $$
    \mathbb{P}\left\{\left|X^{{\tr}} A X-\mathbb{E} X^{\tr} A X\right| \geq t\right\} \leq 2 \exp \left[-c \min \left(\frac{t^{2}}{K^{4}\|A\|_{F}^{2}}, \frac{t}{K^{2}\|A\|}\right)\right],$$
    where $K=\max _{i}\left\|X_{i}\right\|_{\psi_{2}}.$
\end{lemma}

With the above lemma and $\epsilon$-Net argument, we can provide the following result for the norm bound of a vectorized quadratic form.
\begin{lemma}\label{GHWinequality}
    (Norm bound for Vectorized Hanson-Wright Inequality with Gaussian Random Variable)
    Let  $X=\left(X_{1}^{\tr}, \ldots, X_{n}^{\tr}\right)^{\tr}\in \mathbb{R}^{n\times m}$  be a random vector with independent, mean zero Gaussian coordinates, i.e., $X_i\sim\mathcal{N}(0,I_n)$ and $\mathbb{E}\left\{X_i^{\tr} X_j\right\}=O, \forall i\neq j$. Let  $A$  be an  $n \times n$  matrix. Then, for every  $t \geq 0$, we have
    $$
    \mathbb{P}\left\{\left\|X^{\tr} A X\right\|_2^2 \geq 2\operatorname{tr}(A)+t\right\} \leq 2 \exp \left[2m-c \min \left(\frac{t^{2}}{4c^{\prime 4}\|A\|_{F}^{2}}, \frac{t}{2c^{\prime 2}\|A\|}\right)\right],$$
    where $c$ and $c'$ are uniform constant and $\psi_2$ norm of the standard Gaussian random variable, respectively.
\end{lemma}
\begin{proof}
    In this lemma, we use the $\epsilon$-net argument. First, consider the term $\omega^{{\tr}} X^{{\tr}} A X \omega$ for all $\omega$ in unit sphere $S^{m-1}$, note that the vector $\omega^{\tr} X^{\tr}$ is composed with mutually uncorrelated standard Gaussian random variable. With Lemma \ref{HWinequality}, we have
    $$
    \mathbb{P}\left\{ \omega^{\tr} X^{{\tr}} A X \omega - \operatorname{tr}(A) \geq t \right\} \leq 2 \exp \left[-c \min \left(\frac{t^{2}}{c^{\prime 4}\|A\|_{F}^{2}}, \frac{t}{c^{\prime 2}\|A\|}\right)\right].
    $$
    Let $\mathcal{N}_\epsilon$ be the $\epsilon$-net (Definition 
 4.2.1 in \cite{vershynin2018high}) of unit sphere $S^{m-1}$ and take $\epsilon=\frac{1}{2}$, with Corollary 4.2.13 in \cite{vershynin2018high}, there is 
 $\left|\mathcal{N}_\epsilon\right|\leq 5^m$. With Lemma 4.4.1 in \cite{vershynin2018high}, we have
 $$
 \left\|X^{{\tr}} A X\right\|_2^2\leq2\sup_{\omega\in\mathcal{N}_\epsilon}\omega^{\tr} X^{{\tr}} A X \omega.
 $$
Hence there is
$$
\begin{aligned}
    \mathbb{P}\left\{\left\|X^{{\tr}} A X\right\|_2^2 \geq 2\operatorname{tr}(A)+t\right\} \leq& \mathbb{P}\left\{\sup_{\omega\in\mathcal{N}_\epsilon}\omega^{\tr} X^{{\tr}} A X \omega \geq \operatorname{tr}(A)+\frac{t}{2}\right\}\\
    \leq&\sum_{\omega\in\mathcal{N}_\epsilon}\mathbb{P}\left\{\omega^{\tr} X^{{\tr}} A X \omega \geq \operatorname{tr}(A)+\frac{t}{2}\right\}\\
    \leq&2 \exp \left[2m-c \min \left(\frac{t^{2}}{4c^{\prime 4}\|A\|_{F}^{2}}, \frac{t}{2c^{\prime 2}\|A\|}\right)\right],
\end{aligned}
$$
where the last inequality holds since $5<e^2$.
\end{proof}

\section{Proof of \Cref{thm1}}\label{secMainProof}

We here complete all the proof details of \Cref{thm1}. 

\subsection{General Analysis}

Denote $\tilde{Z}_{k,p}=D_{p} Z_{k, p}$ as the modified sample at time step $k$. 
Note that
$$\begin{aligned}
\tilde{y}_{k+1}= & \left(\sum_{l=p}^{k}\left(G_{p} Z_{l, p}+b_{l, p}+e_{l}\right) Z_{l, p}^{{\tr}}\right) \bar{V}_{k, p}^{-1} Z_{k+1, p} \\
= & G_{p}\left(I-\lambda D_{p}^{-2} \bar{V}_{k, p}^{-1}\right) Z_{k+1, p}+\sum_{l=p}^{k} b_{l, p} Z_{l, p}^{{\tr}} \bar{V}_{k, p}^{-1} Z_{k+1, p}  +\sum_{l=p}^{k} e_{l, p} Z_{l, p}^{{\tr}} \bar{V}_{k, p}^{-1} Z_{k+1, p}.
\end{aligned}$$
Hence we have
$$
\begin{aligned}
\tilde{y}_{k+1}-\hat{y}_{k+1}= & -\lambda G_{p} D_{p}^{-2} \bar{V}_{k, p}^{-1} Z_{k+1, p}+\sum_{l=p}^{k} b_{l, p} Z_{l, p}^{{\tr}} \bar{V}_{k, p}^{-1} Z_{k+1, p} +\sum_{l=p}^{k} e_{l} Z_{l, p}^{{\tr}} \bar{V}_{k, p}^{-1} Z_{k+1, p}-b_{k+1, p}.
\end{aligned}
$$
The difference at time step $k+1$ can be formulated as
$$
\begin{aligned}
\left\|\tilde{y}_{k+1}-\hat{y}_{k+1}\right\|_{2}^{2} \leq & 6\left(\left\| \lambda G_{p} D_{p}^{-2} \bar{V}_{k, p}^{-\frac{1}{2}}\right\|_{2}^{2}+\left\| B_{k, p} \bar{Z}_{k, p}^{{\tr}} \bar{V}_{k, p}^{-\frac{1}{2}} \right\|_{2}^{2} +\left\|E_{k, p} \bar{Z}_{k, p}^{{\tr}} \bar{V}_{k, p}^{-\frac{1}{2}}\right\|_{2}^{2}\right)\left\|\bar{V}_{k, p}^{-\frac{1}{2}} Z_{k+1,p}\right\|_{2}^{2}+2\left\|b_{k+1, p}\right\|_{2}^{2}.
\end{aligned}$$
where $b_{k,p}=C\left(A-LC\right)^p \hat{x}_{k-p}$ denotes the bias induced by truncation. Due to the fact that the parameter $p$ varies with the time step $k$, we also need to divide the gap term $\mathcal{L}_N$ into different epochs. Denote the number of all the epoch is $N_E$, i.e., $N=2^{N_E}T_{\text{init}}$, and
let $T_l = 2^{l-1}T_{\text{init}}+1$ be the beginning time step of the $l$-th epoch, then we have
$$
\begin{aligned}
\mathcal{L}_{N}=&\sum_{l=1}^{N_E}\sum_{k=T_l}^{2T_l-2}\left\|\tilde{y}_{k+1}-\hat{y}_{k+1}\right\|_{2}^{2}\\
\leq&\sum_{l=1}^{N_E}\sup_{T_l\leq k\leq 2T_l-2
}6\left(\left\| \lambda G_{p} D_{p}^{-2} \bar{V}_{k, p}^{-\frac{1}{2}}\right\|_{2}^{2}+\left\| B_{k, p} \bar{Z}_{k, p}^{{\tr}} \bar{V}_{k, p}^{-\frac{1}{2}} \right\|_{2}^{2} +\left\|E_{k, p} \bar{Z}_{k, p}^{{\tr}} \bar{V}_{k, p}^{-\frac{1}{2}}\right\|_{2}^{2}\right)\\
&\times\sum_{k=T_l}^{2T_l-1}\left\|\bar{V}_{k, p}^{-\frac{1}{2}} Z_{k+1,p}\right\|_{2}^{2}+2\sum_{l=1}^{N_E}\sum_{k=T_l}^{2T_l-2}\left\|b_{k+1, p}\right\|_{2}^{2}.\\
\end{aligned}
$$
Moreover, for the term $\sum_{k=T_l}^{2T_l-2}\left\|\bar{V}_{k, p}^{-\frac{1}{2}} Z_{k+1,p}\right\|_{2}^{2}$, we further have
$$
\sum_{k=T_l}^{2T_l-2}\left\|\bar{V}_{k, p}^{-\frac{1}{2}} Z_{k+1,p}\right\|_{2}^{2}\leq \sup_{T_l\leq k\leq 2T_l-2
}\left\|\bar{V}_{k, p}^{-\frac{1}{2}} \bar{V}_{k+1, p}^{\frac{1}{2}}\right\|_{2}^{2}\sum_{k=T_l}^{2T_l-2}\left\|\bar{V}_{k+1, p}^{-\frac{1}{2}} Z_{k+1,p}\right\|_{2}^{2}.
$$
In the following analysis, we will gradually analyze the above terms from a new perspective and then prove the logarithm bound provided in Theorem 1 with high probability $1-\delta$.

\subsection{Uniform Boundedness of Bias error and the selection of parameter $\beta$}\label{sectionBeta}

Note that the bias error are in two parts, i.e., the bias regression error $\left\| B_{k, p} \bar{Z}_{k, p}^{{\tr}} \bar{V}_{k, p}^{-\frac{1}{2}} \right\|_{2}^{2}$ and bias accumulation error $\sum_{k=T_l}^{2T_l-1}\left\|b_{k+1, p}\right\|_{2}^{2}$ for each epoch. We first consider the first term, note that at each epoch $l$, for any $T_l\leq k\leq 2T_l-2$ there is
$$
\begin{aligned}
    \left\| B_{k, p} \bar{Z}_{k, p}^{{\tr}} \bar{V}_{k, p}^{-\frac{1}{2}} \right\|_{2}^{2}=& \left\|B_{k, p} \bar{Z}_{k, p}^{{\tr}} \bar{V}_{k, p}^{-1}\bar{Z}_{k, p} B_{k,p}^{\tr}\right\|_2^2
    \leq \sum_{i=p}^{k}\left\|b_{i, p}\right\|_{2}^{2}\leq\sum_{i=p}^{k}\left\|C(A-LC)^p\right\|_2^2\left\|\hat{x}_{i-p}\right\|_2^2.
\end{aligned}
$$
where $p=\beta \log(T_l-1)$.
Conditioned on the event $\mathcal{E}_{x}$, for a fixed probability $\delta_1$, we have that for all $k\in\mathbb{N}$, there is
$$
\left\|\hat{x}_{k}\right\|_2^2\leq \left(\sqrt{n}+\sqrt{2 \log \frac{4k}{\delta_1}}\right)^2\left\|\Gamma_k\right\|_2^2\leq \left(\sqrt{n}+\sqrt{2 \log \frac{4k}{\delta_1}}\right)^2\left\|Q\right\|_2\sum_{i=0}^{k-1}\left\|A^i\right\|_2^2.
$$
Hence there exists a number $M_1=\operatorname{poly}\left(M,n,\log\left(\frac{1}{\delta_1}\right)\right)$ only related to system parameter and $\delta_1$, such that 
$$
\left\|\hat{x}_{k}\right\|_2^2 \leq M_1 k^{2\kappa-1}\log(4k),\;\; \forall k\ge T_{\text{init}}.
$$
Note that there is $\left\|C(A-LC)^p\right\|_2^2\leq M\rho(A-LC)^p$. 
We choose $\beta=\frac{M_3}{\log(1/\rho(A-LC))}$, then we have
$$
\begin{aligned}
    \left\| B_{k, p} \bar{Z}_{k, p}^{{\tr}} \bar{V}_{k, p}^{-\frac{1}{2}} \right\|_{2}^{2}\leq& MM_1(2T_l)^{2\kappa+1}\rho(A-LC)^{\frac{M3}{\log(1/\rho(A-LC))}\log(T_l-1)}\\
    \leq&\operatorname{poly}\left(M,M_1,n,\log\left(\frac{1}{\delta_1}\right)\right)(2T_l)^{2\kappa+1}\frac{1}{(T_l-1)^{M3}}.
\end{aligned}
$$
Note that $M_1$ is only related to system parameters, and we only need to choose $M_3=2\kappa+1$, then there is
$$
\left\| B_{k, p} \bar{Z}_{k, p}^{{\tr}} \bar{V}_{k, p}^{-\frac{1}{2}} \right\|_{2}^{2}\leq\sum_{i=p}^{2T_l-2}\left\|b_{i, p}\right\|_{2}^{2}\leq\operatorname{poly}\left(M,n,\log\left(\frac{1}{\delta_1}\right)\right).
$$
Note that the above bound holds for all $T_l\leq k\leq 2T_l-2$, together with the arbitrariness of $l$, we have 
$$
\sup_{1\leq l\leq N_E}\sup_{T_l\leq k\leq 2T_l-2}\left\| B_{k, p} \bar{Z}_{k, p}^{{\tr}} \bar{V}_{k, p}^{-\frac{1}{2}} \right\|_{2}^{2}\leq\operatorname{poly}\left(M,\log\left(\frac{1}{\delta_1}\right)\right).
$$
Moreover, for the term  $\sum_{l=1}^{N_E}\sum_{k=T_l}^{2T_l-2}\left\|b_{k+1, p}\right\|_{2}^{2}$, we have
$$
\sum_{l=1}^{N_E}\sum_{k=T_l}^{2T_l-2}\left\|b_{k+1, p}\right\|_{2}^{2}\leq \frac{\log(N/T_{\text{init}})}{\log2} \operatorname{poly}\left(M,\log\left(\frac{1}{\delta_1}\right)\right)=O(\log N).
$$
Therefore it is sufficient to guarantee the uniform boundedness of bias error only with the parameter $\beta$ chosen to be proportional to $1/\log\rho(A-LC)$ and the order $\kappa$ of the marginal stable Jordan block of $A$, which is independent to the time step $k$ or the initial time $T_{\text{init}}$. 

\subsection{Persistent Excitation of the Gram matrix $\bar{Z}_{k,p}\bar{Z}_{k,p}^{\tr}$ for all $k\ge T_{\text{init}}$}
Consider the event
$$
\mathcal{E}_{PE} \triangleq\left\{\bar{Z}_{k,p}\bar{Z}_{k,p}^{\tr}\ge \frac{\sigma_R}{4}kI,\;\; \forall k\ge T_{\text{init}}\right\}.$$
From Lemma \ref{persistent}, we can obtain that for the above fixed $\beta$ and $\delta_1$, there exists a number $N_0=\allowbreak\operatorname{poly}\Big(M,\beta,m, \allowbreak \log\left(\frac{1}{\delta_1}\right)\Big),$ such that the event $\mathcal{E}_{PE}$ holds for all $k>N_0$ with probability at least $1-\delta_1$. Without loss of generality, we can assume that the length of the warm-up horizon $T_{\text{init}}$ satisfies $T_{\text{init}}>N_0$. Then the persistent excitation condition holds for all $k\ge T_{\text{init}}$ with probability $1-\delta_1$. This condition will be repeatedly used in the following analysis.

\subsection{Uniform boundedness of $\left\| \lambda G_{p} D_{p}^{-2} \bar{V}_{k, p}^{-\frac{1}{2}}\right\|_{2}^{2}$ with high probability}
For the regularization error term $\left\| \lambda G_{p} D_{p}^{-2} \bar{V}_{k, p}^{-\frac{1}{2}}\right\|_{2}^{2}$, we have
$$
\begin{aligned}
    \lambda G_{p} D_{p}^{-2} \bar{V}_{k, p}^{-\frac{1}{2}}*\left(\lambda G_{p} D_{p}^{-2} \bar{V}_{k, p}^{-\frac{1}{2}}\right)^{\tr}=&\lambda^2G_pD_p^{-1}\tilde{V}_{k,p}^{-1}D_p^{-1}G_p^{\tr}\leq \lambda G_pD_p^{-1}D_p^{-1}G_p^{\tr}\\=&\sum_{l=0}^{p-1}\frac{1}{\gamma^{2l}}C(A-LC)^l LL^{\tr} (A-LC)^{l{\tr}}C^{\tr}\\
    \leq&\left\|L\right\|_2^2\left\|C\right\|_2^2\sum_{l=0}^{p-1}\frac{1}{\gamma^{2l}}\left\|(A-LC)^l\right\|_2^2.
\end{aligned}
$$
From the basic assumption of $A-LC$, i.e., $\left\|(A-LC)^p\right\|_2\leq M\rho(A-LC)^p$, we have 
$$
\left\| \lambda G_{p} D_{p}^{-2} \bar{V}_{k, p}^{-\frac{1}{2}}\right\|_{2}^{2}\leq \operatorname{poly}(M)\frac{1}{1-\left(\frac{\rho(A-LC)}{\gamma}\right)^2}, \;\;\forall k\ge T_{\text{init}}.
$$
\subsection{Uniform Boundedness of $\left\|E_{k, p} \bar{Z}_{k, p}^{{\tr}} \bar{V}_{k, p}^{-\frac{1}{2}}\right\|_{2}^{2}$ with high probability}

In this subsection, we will provide a uniform upper bound of the regression error term  $\left\|E_{k, p} \bar{Z}_{k, p}^{{\tr}} \bar{V}_{k, p}^{-\frac{1}{2}}\right\|_{2}^{2}$ for all $k\ge T_{\text{init}}$.
Note that $e_k$ are mutually uncorrelated Gaussian random variable, i.e., $e_k\sim \mathcal{N}(0,\bar{R})$, hence the matrix  
$\bar{R}^{-\frac{1}{2}}E_{k,p}=\begin{bmatrix}
   \bar{R}^{-\frac{1}{2}}e_p&\cdots& \bar{R}^{-\frac{1}{2}}e_k
\end{bmatrix}$ is composed of mutually uncorrelated standard Gaussian random vectors. Then we consider the matrix $$\mathcal{E}_{k,p}\triangleq\bar{R}^{-\frac{1}{2}}E_{k,p}\bar{Z}_{k, p}^{{\tr}} \bar{V}_{k, p}^{-1}\bar{Z}_{k, p}E_{k,p}^{{\tr}}\bar{R}^{-\frac{1}{2}},$$
and denote $V_{k,p}^{Z}=\bar{Z}_{k, p}^{{\tr}} \bar{V}_{k, p}^{-1}\bar{Z}_{k, p}$. Then
with Lemma \ref{GHWinequality}, we have

$$
\mathbb{P}\left\{\left\|\mathcal{E}_{k,p}\right\|_2^2 \geq 2\operatorname{tr}(V_{k,p}^Z)+t\right\} \leq 2 \exp \left[2m-c \min \left(\frac{t^{2}}{4c^{\prime 4}\|V_{k,p}^Z\|_{F}^{2}}, \frac{t}{2c^{\prime 2}\|V_{k,p}^Z\|_2}\right)\right],$$
where $c$ are global constants related to $\psi_2$-norm and $c'$ is the $\psi_2$ norm of standard Gaussian random variable.
From the fact that 
$
    V_{k,p}^{Z}=\bar{Z}_{k, p}^{{\tr}} \left(\lambda D_p^{-2}+\bar{Z}_{k, p}\bar{Z}_{k, p}^{{\tr}}\right)^{-1}\bar{Z}_{k, p},
$
we have
$$
\begin{aligned}
    \text{tr}(V_{k,p}^{Z})=&\text{tr}\left( \left(\lambda D_p^{-2}+\bar{Z}_{k, p}\bar{Z}_{k, p}^{{\tr}}\right)^{-1}\bar{Z}_{k, p}\bar{Z}_{k, p}^{{\tr}}\right)=mp-\lambda\text{tr}(\tilde{V}_{k,p}^{-1}),\\
    \left\|V_{k,p}^{Z}\right\|_F^2=&\text{tr}\left(V_{k,p}^{Z}(V_{k,p}^{Z})^{\tr}\right)=\text{tr}\left(\left(I-\lambda\tilde{V}_{k,p}^{-1}\right)^2\right)\leq mp,\\
\end{aligned}
$$
and $\left\|V_{k,p}^{Z}\right\|_2\leq 1$.

Then we take $$t=\max\left\{\frac{2c^{\prime 2}}{\sqrt{c}}mp,\frac{2c^{\prime 2}}{c}\left(\log\left(\frac{k^2}{\delta_1}\right)+2m\right)\right\},$$
and we can verify that 
$$
c\min \left\{\frac{t^{2}}{4c^{\prime 4}\|V_{k,p}^Z\|_{F}^{2}}, \frac{t}{2c^{\prime 2}\|V_{k,p}^Z\|_2}\right\}\ge \max\left\{\log\left(\frac{2k^2}{\delta_1}\right)+2m,m\beta \log k\right\}.
$$
Hence for each $k\ge T_{\text{init}}$, we have
$$
\mathbb{P}\left\{\left\|\mathcal{E}_{k,p}\right\|_2 \geq 2\operatorname{tr}(V_{k,p}^Z)+t\right\} \leq \frac{\delta_1}{k^2}.
$$
Take a uniform bound over all $k$, and note that $p\leq \beta \log(k)$, we have
$$
\mathbb{P}\left\{\left\|\mathcal{E}_{k,p}\right\|_2 \leq \operatorname{poly}\left((m,\beta,\log\left(\frac{1}{\delta_1}\right)\right) \log k,\;\;\forall k\ge T_{\text{init}}\right\}\ge 1-\frac{\pi^2\delta_1}{6},
$$
and
$$
\mathbb{P}\left\{\left\|E_{k, p} \bar{Z}_{k, p}^{{\tr}} \bar{V}_{k, p}^{-\frac{1}{2}}\right\|_{2}^2 \leq \bar{\sigma}_{\bar{R}}\operatorname{poly}\left((m,\beta,\log\left(\frac{1}{\delta_1}\right)\right) \log k,\;\;\forall k\ge T_{\text{init}}\right\}\ge 1-\frac{\pi^2\delta_1}{6},
$$
where $\bar{\sigma}_{\bar{R}}$ is the largest singular value of $\bar{R}$.

\subsection{Uniform Boundedness of $\left\|\bar{V}_{k,p}^{-\frac{1}{2}}\bar{V}_{k+1,p}^{\frac{1}{2}}\right\|_2^2$ with high probability}
Inspired by \cite{tsiamis9894660}, we still consider the successive representation of $Z_{k,p}$ generated by the minimal polynomial of $A$.
For the matrix $A$, suppose the minimal polynomial of $A$ takes the form as
$$
A^d + a_{d-1}A^{d-1}+\dots+a_0=0,
$$
where $d$ is the dimension of the minimal polynomial of $A$.
Then from the Lemma 2 of \cite{tsiamis9894660}, we have the following successive representation of the sample $Z_{k,p}$
$$Z_{k,p}=a_{d-1} Z_{k-1,p}+\ldots+a_{0} Z_{k-d,p}+\sum_{s=0}^{d} \operatorname{diag}\left(L_{s}, \ldots, L_{s}\right) \tilde{E}_{k-s,p},$$
where $L_{s}=-a_{d-s} I_{m}-\sum_{t=1}^{s-1} a_{d-s+t} C A^{t-1} L+C A^{s-1} L$ and $\tilde{E}_{k,p}=[e_{k-p}^\tr,\dots,e_{k-1}^\tr]^\tr$. We also have $$\left\|L_{s}\right\|_{2} \leq\|a\|_{1}\|C\|_{2}\|L\|_{2} \max _{0 \leq i \leq d}\left\|A^{i-1}\right\|_{2}.$$
Denote $\delta_{k} \triangleq \sum_{s=0}^{d} \operatorname{diag}\left(L_{s}, \ldots, L_{s}\right) \tilde{E}_{k-s,p}$, then there is 
$$
\left\|\delta_{k}\right\|_{2} \leq \Delta\sup_{0\leq s\leq d+p}\left\|e_{k-s}\right\|_{2},\qquad
\Delta \triangleq(d+1) \max _{0 \leq s \leq d}\left\|L_{s}\right\|_{2} \sqrt{p}.$$
Then we consider the term $\left\|\bar{V}_{k,p}^{-\frac{1}{2}}\bar{V}_{k+1,p}^{\frac{1}{2}}\right\|_2^2$, together with the structure of $\bar{V}_{k,p}$, we have
$$
\begin{aligned}
    \left\|\bar{V}_{k,p}^{-\frac{1}{2}}\bar{V}_{k+1,p}^{\frac{1}{2}}\right\|_2^2=&1+Z_{k+1,p}\bar{V}_{k,p}^{-1}Z_{k+1,p} \\
    \leq& 1+2\delta_{k+1}^{\tr} \bar{V}_{k,p}^{-1}\delta_{k+1}+2d\left\|a\right\|_2^2\sum_{s = 0}^{d-1}Z_{k-s,p}^{\tr} \bar{V}_{k,p}^{-1}Z_{k-s,p}.
\end{aligned}
$$
Conditioned on $\mathcal{E}_e$ and $\mathcal{E}_{PE}$, we have
$$
\begin{aligned}
    \delta_k^{\tr} \bar{V}_{k,p}^{-1}\delta_k\leq& \frac{\Delta^2}{\lambda+\frac{\sigma_R}{4}k}
\sup_{0\leq s\leq d+p}\left\|e_{k-s}\right\|_{2}^2
\leq \frac{M(d+1)^2 \beta\operatorname{log}(2T_l)}{\lambda+\frac{\sigma_R}{4}T_l}\left(\sqrt{m}+\sqrt{2 \log \frac{2 T_l}{\delta_{1}}}\right)^2\bar{\sigma}_{\bar{R}}.
\end{aligned}
$$
The above inequality holds due to the fact that for any $k\ge T_{\text{init}}$, there exists $l\leq N_E$ such that $T_l\leq k\leq 2T_l-2$.
Note that the parameters $d,\beta,M,m$ are not related to the time step $k$, hence we have 
$$
\begin{aligned}
    \delta_k^{\tr} \bar{V}_{k,p}^{-1}\delta_k\leq&\frac{\operatorname{poly}\left(\frac{d}{M\sigma_R},\beta,\operatorname{log}\frac{1}{\delta_1},\bar{\sigma}_{\bar{R}}\right) \operatorname{log}^2(2T_l)}{T_l}
    \leq \frac{8}{e^2}\operatorname{poly}\left(M,\frac{d}{\sigma_R},\beta,\operatorname{log}\frac{1}{\delta_1},\bar{\sigma}_{\bar{R}}\right).
\end{aligned}
$$

For the quadratic form $\sum_{s = 0}^{d-1}Z_{k-s,p}^{\tr} \bar{V}_{k,p}^{-1}Z_{k-s,p}$,
from Woodbury Equality, there is
$$
\begin{aligned}
    Z_{k,p}^{\tr}\left(V_{k-1, p}+Z_{k, p} Z_{k, p}^{{\tr}}\right)^{-1}Z_{k,p}=&Z_{k,p}^{\tr} V_{k-1, p}^{-1}Z_{k,p}-\frac{\left(Z_{k,p}^{\tr} V_{k-1, p}^{-1} Z_{k, p}\right)^2 }{1+Z_{k, p}^{{\tr}} V_{k-1, p}^{-1} Z_{k, p}}=\frac{Z_{k, p}^{{\tr}}V_{k-1,p}^{-1}Z_{k, p}}{1+Z_{k, p}^{{\tr}}V_{k-1,p}^{-1}Z_{k, p}}.
\end{aligned}
$$
With a similar technique, it is easy to verify that 
$$
2d\left\|a\right\|_2^2\sum_{s = 0}^{d-1}Z_{k-s,p}^{\tr} \bar{V}_{k,p}^{-1}Z_{k-s,p}\leq d^2\left\|a\right\|_2^2.
$$
Then we have that conditioned on event $\mathcal{E}_e$ and $\mathcal{E}_{PE}$, we can find a bound $M_1=\operatorname{poly}\left(M,\bar{\sigma}_{\bar{R}}, \frac{d}{\sigma_R},\left\|a\right\|_2,\beta,\operatorname{log}\frac{1}{\delta_1}\right)$ for the term $\left\|\bar{V}_{k,p}^{-\frac{1}{2}}\bar{V}_{k+1,p}^{\frac{1}{2}}\right\|_2^2$ with high probability $1-\left(1+\frac{\pi^2}{6}\right)\delta_1$.
\subsection{Uniform Boundedness of $\sum_{k=T_{\text{init}}}^{N}\left\|\bar{V}_{k, p}^{-\frac{1}{2}} Z_{k,p}\right\|_{2}^{2}$ with high probability}
We first consider the accumulation error at the $l$-th epoch, where $T_l = 2^{l-1}T_{\text{init}}+1$. We first have the following result for the accumulation error
$$
\sum_{k=T_{l}}^{2T_l-2}\left\|\bar{V}_{k+1, p}^{-\frac{1}{2}} Z_{k+1,p}\right\|_{2}^{2}\leq \log\frac{\det(\bar{V}_{2T_l-1,p})}{\det(\bar{V}_{T_l,p})}.
$$
Moreover, consider the expression of $\bar{V}_{2T_l-2,p}$, we also have
$$
\frac{\det(\bar{V}_{2T_l-1,p})}{\det(\bar{V}_{T_l,p})}=\frac{\det(\bar{V}_{2T_l-1,p})\det(D_p^{-2})}{\det(\bar{V}_{T_l,p})\det(D_p^{-2})}=\frac{\det(\tilde{V}_{2T_l-1,p})}{\det(\tilde{V}_{T_l,p})}.
$$
Then we consider the accumulation error of two successive epochs, i.e.,
$$
\sum_{k=T_{l}}^{4T_l-4}\left\|\bar{V}_{k+1, p}^{-\frac{1}{2}} Z_{k+1,p}\right\|_{2}^{2}\leq \log\frac{\det(\tilde{V}_{4T_l-3,p^\prime})}{\det(\tilde{V}_{2T_l-1,p^\prime})}+\log\frac{\det(\tilde{V}_{2T_l-1,p})}{\det(\tilde{V}_{T_l,p})},
$$
where $p'=\beta \log(2T_l-2)$ and $p=\beta \log(T_l-1)$, hence we have $p'-p\leq \beta \log 2$.
Note that we can separate the augmented samples $Z_{k,p^\prime}$and Gram matrices $\tilde{V}_{k,p^{\prime}}$as   
$$
Z_{k,p'} = \begin{bmatrix}
   Z_{k-p,p^\prime-p}^{{\tr}}&Z_{k,p}^{\tr}
\end{bmatrix}^{\tr},$$
and
$$
\begin{aligned}
\tilde{V}_{2T_l-1,p'}=&\lambda I + D_{p^{\prime}}\bar{Z}_{2T_l-1,p'}\bar{Z}_{2T_l-1,p'}^{\tr} D_{p^{\prime}}\\
=&\lambda I + \sum_{k=p'}^{2T_l-1}D_{p^{\prime}}Z_{k,p'}Z_{k,p'}^{\tr} D_{p^{\prime}}\\
=&\lambda I +D_{p^{\prime}}\begin{bmatrix}
Z_{11}&Z_{12}\\Z_{12}^{\tr}&Z_{22}
\end{bmatrix} D_{p^{\prime}},
\end{aligned} 
$$
where 
$
    Z_{11} = \sum_{k=p'}^{2T_l-1}Z_{k-p,p^\prime-p}Z_{k-p,p^\prime-p}^{{\tr}}$,$
    Z_{12} = \sum_{k=p'}^{2T_l-1}Z_{k-p,p^\prime-p}Z_{k,p}^{{\tr}}
$ and $Z_{22} = \sum_{k=p'}^{2T_l-1}Z_{k,p}Z_{k,p}^{{\tr}}$.

Conditioned on event $\mathcal{E}_{PE}$, we have
$$
Z_{22}\ge \frac{\sigma_R}{4}(2T_l-1)I.
$$
This is because the smallest eigenvalue of a positive definite matrix is smaller than that of any  diagonal sub-matrix, i.e., $\lambda_{\min}(Z_{22})>\lambda_{\min}\left(\begin{bmatrix}
Z_{11}&Z_{12}\\Z_{12}^{\tr}&Z_{22}\end{bmatrix}\right).$ 
Split the matrix $D_{p^\prime}$ into two parts, i.e., $D_{p^\prime}=\operatorname{diag}(D_{p^{\prime}/p},D_p)$, where $D_{p^{\prime}/p}=\operatorname{diag}(\gamma^{p^{\prime}},\dots,\gamma^{p+1})$. Then we have
$$
\begin{aligned}
\tilde{V}_{2T_l-1,p'}=&
\begin{bmatrix}
\lambda I +D_{p^{\prime}/p}Z_{11}D_{p^{\prime}/p}&D_{p^{\prime}/p}Z_{12}D_p\\D_p Z_{12}^{\tr} D_{p^{\prime}/p}&\lambda I+D_pZ_{22}D_p
\end{bmatrix}\\
=& \begin{bmatrix}
    \lambda I +D_{p^{\prime}/p}Z_{11}D_{p^{\prime}/p}&D_{p^{\prime}/p}Z_{12}D_p\\D_p Z_{12}^{\tr} D_{p^{\prime}/p}&\tilde{V}_{2T_l-1,p}-\sum_{k=p}^{p^{\prime}-1}D_p Z_{k,p}Z_{k,p}^{\tr} D_p
\end{bmatrix}
\\ \triangleq&
\begin{bmatrix}
    \tilde{Z}_{11}&\tilde{Z}_{12}\\
    \tilde{Z}_{12}^{\tr}& \tilde{Z}_{22}
\end{bmatrix}.
\end{aligned}
$$
where the last notation is for simplification.
Note that 
$$
\begin{aligned}
    \det\tilde{V}_{2T_l-1,p'}=&\det\begin{bmatrix}
    I&-\tilde{Z}_{12}\tilde{Z}_{22}^{-1}\\
    0&I
\end{bmatrix}\begin{bmatrix}
    \tilde{Z}_{11}&\tilde{Z}_{12}\\
    \tilde{Z}_{12}^{\tr}& \tilde{Z}_{22}
\end{bmatrix}\begin{bmatrix}
    I&\\
    -\tilde{Z}_{22}^{-1}\tilde{Z}_{12}^{{\tr}}&I
\end{bmatrix}\\
=&\det\tilde{Z}_{22} \cdot
\det(\tilde{Z}_{11}-\tilde{Z}_{12}\tilde{Z}_{22}^{-1}\tilde{Z}_{12}^{\tr}).
\end{aligned}
$$
Moreover, consider the vector $\tilde{v}=\begin{bmatrix}
    v^{\tr} & v^{\tr} \tilde{Z}_{12} \tilde{Z}_{22}^{-1}
\end{bmatrix}^{\tr}$, where $v$ is an arbitrary vector. Then
there is
$$
\tilde{v}^{\tr} \tilde{V}_{2T_l-1,p'} \tilde{v}=v^{\tr}\left(\tilde{Z}_{11}-\tilde{Z}_{12}\tilde{Z}_{22}^{-1}\tilde{Z}_{12}^{\tr}\right)v^{\tr}\ge \lambda \tilde{v}^{\tr} \tilde{v}\ge \lambda v^{\tr} v.
$$
Due to the arbitrariness of $v$, we have
$$
\det(\tilde{Z}_{11}-\tilde{Z}_{12}\tilde{Z}_{22}^{-1}\tilde{Z}_{12}^{\tr})\ge \det(\lambda I)\ge \lambda^{m(p^\prime - p)}.$$
While for the term 
$$
\det\tilde{Z}_{22}=\det\left(\tilde{V}_{2T_l-2}-\sum_{k=p}^{p^{\prime}-1}D_p Z_{k,p}Z_{k,p}^{\tr} D_p\right),
$$
conditioned on the event $\mathcal{E}_{PE}$ and $\mathcal{E}_{Z}$, we have 
$$
\left\|\left(\Gamma_{k,p}^Z\right)^{-1 / 2} Z_{k,p}\right\|_{2} \leq \sqrt{mp}+\sqrt{2 \log \frac{4 k}{\delta_1}}.$$
Hence, we can obtain that
$$
\begin{aligned}
    Z_{k,p}Z_{k,p}^{\tr} \leq& \left(\sqrt{mp}+\sqrt{2 \log \frac{4 k}{\delta_1}}\right)^2\Gamma_{k,p}^Z\\
    \leq&\left(2mp+4\log\left(\frac{4k}{\delta_1}\right)\right)\mathbb{E}\left\{Z_{k,p}^TZ_{k,p}\right\} I_{mp}\\
    \leq& M\left(2mp+4\log\left(\frac{4k}{\delta_1}\right)\right)p k^{2\kappa-1}I_{mp}.\\
\end{aligned}
$$
We have
$$
\sum_{k=p}^{p^{\prime}-1} Z_{k,p}Z_{k,p}^{\tr} \leq (p^\prime-p)M\left(2mp+4\log\left(\frac{4p^{\prime}}{\delta_1}\right)\right)p^{\prime2\kappa}I_{mp}.
$$
Conditioned on $\mathcal{E}_{PE}$, we have
$$
\sum_{k=p}^{p^{\prime}-1} Z_{k,p}Z_{k,p}^{\tr} \leq (p^\prime-p)M\frac{4\left(2mp+4\log(\frac{4p^\prime}{\delta_1})\right)p^{\prime2\kappa}}{\sigma_R(2T_l-1)}Z_{22}.
$$
Note that $p^{\prime}=\beta\log(2T_l-2)$, and $p=\beta\log(T_l-1)$ hence there exists a uniform $M_1$, such that
$$
M_1\ge\frac{4\left(2mp+4\log(\frac{4p^\prime}{\delta_1})\right)(p^{\prime})^{2\kappa}}{\sigma_R(2T_l-1)},\;\;\forall l=1,\dots,N_E,\dots
$$
and we can verify that
$$
M_1=\operatorname{poly}\left(\frac{1}{\sigma_R},m,\beta,\log\left(\frac{1}{\delta_1}\right)\right).$$
Then we have
$$
D_p\bar{Z}_{2T_l-2,p}\bar{Z}_{2T_l-2,p}^{\tr} D_p\leq(1+M_1)D_pZ_{22}D_p,
$$
and
$$
\tilde{V}_{2T_l-2,p}\leq (1+M_1)\tilde{Z}_{22}.
$$
Finally, we can show that for all $l=1,\dots,N_E$
$$
\begin{aligned}
    &\log\det\tilde{V}_{2T_l-2,p}-\log\det\tilde{V}_{2T_l-2,p^\prime}\\
    =&\log\det\tilde{V}_{2T_l-2,p}-(\log\det\tilde{Z}_{22}+\log\det(\tilde{Z}_{11}-\tilde{Z}_{12}\tilde{Z}_{22}^{-1}\tilde{Z}_{12}^{\tr}))\\\leq& mp\log(1+M_1)-m(p^{\prime}-p)\log(\lambda).
\end{aligned}
$$

We denote $p_l=\beta\log(2^{l-1}T_{\text{init}})$, then we have
$$
\begin{aligned} \sum_{k=T_{\text{init}}}^{N}\left\|\bar{V}_{k, p}^{-\frac{1}{2}} Z_{k,p}\right\|_{2}^{2}=&\sum_{l=1}^{N_E}\log\frac{\det(\tilde{V}_{2T_l-1,p_l})}{\det(\tilde{V}_{T_l,p_l})}\\
\leq& \log\det(\tilde{V}_{2T_{N_E}-2,p_{N_E}})-\log\det(\tilde{V}_{T_{1}-1,p_{1}})\\&+\sum_{l=1}^{N_E}(1+M_1)mp_l-N_E\beta\log(2)\log(\lambda)\\
\leq &\log\det(\tilde{V}_{N,p_{N_E}})+\log(1+M_1)N_E m\beta\log(N).
\end{aligned}
$$
Note that $N=2^{N_E}T_{\text{init}}$, conditioned on the event $\mathcal{E}_{Z}$, we have
$$
\det(\tilde{V}_{N,p_{N_E}})\leq \det\left(\operatorname{poly}\left(M,m,\beta,\log\frac{1}{\delta_1}\right)N^{2\kappa} I_{mp_{N_E}}\right)\leq \left(\operatorname{poly}\left(M,m,\beta,\log\frac{1}{\delta_1}\right)N^{2\kappa} I_{mp_{N_E}}\right)^{(m\beta \log(N/2))}.
$$
Hence we have $N_E=\frac{\log(N/T_{\text{init}})}{\log2}$, and
$$
\sum_{k=T_{\text{init}}}^{N}\left\|\bar{V}_{k, p}^{-\frac{1}{2}} Z_{k,p}\right\|_{2}^{2}=\operatorname{poly}\left(M,\frac{1}{\sigma_R},m,\beta,\log\frac{1}{\delta_1}\right)O\left(\log^2(N)\right).
$$

\subsection{Uniform Boundedness of $\left(\hat{y}-\Tilde{y}\right)^{\tr} e_k$}
In this subsection, in order to simplify the proof, we directly utilize the result in Theorem 1 and Theorem 3 in \cite{tsiamis9894660}, and we have
$$
\sum_{k=T_{\text{init}}}^{N}e_k^\top(\hat{y}_k-\tilde{y}_k)=\operatorname{poly}\left(\frac{1}{\delta_1}\right)\sqrt{\mathcal{L}_N}\log(\mathcal{L}_N)=\operatorname{poly}\left(\frac{1}{\delta_1}\right)o(\mathcal{L}_N)
$$
with probability $1-\delta_{1}$.

\subsection{Uniform logarithm bound for the whole regret}
Based on the above analysis, we have the following facts collected:
for a given probability $\delta_1$, $M$ is a constant only related to system parameters $A,C,Q,R$, then we have the following facts 
\begin{enumerate}
    \item If $\beta$ is chosen to be $\beta=\frac{2\kappa+1}{\log(1/\rho(A-LC))}$, then conditioned on $\mathcal{E}_x$ we have
$$
\mathbb{P}\left\{
\sup_{1\leq l\leq N_E}\sup_{T_l\leq k\leq 2T_l-2}\left\| B_{k, p} \bar{Z}_{k, p}^{{\tr}} \bar{V}_{k, p}^{-\frac{1}{2}} \right\|_{2}^{2}\leq\operatorname{poly}\left(M,\log\left(\frac{1}{\delta_1}\right)\right)\right\}\ge 1-\frac{\pi^2\delta_1}{6},
$$
and
$$
\mathbb{P}\left\{\sum_{l=1}^{N_E}\sum_{k=T_l}^{2T_l-2}\left\|b_{k+1, p}\right\|_{2}^{2}\leq \operatorname{poly}\left(M,\log\left(\frac{1}{\delta_1}\right)\right)\log N\right\}\ge 1-\frac{\pi^2\delta_1}{6}.
$$
    \item Conditioned on $\mathcal{E}_e$, we have
    $$
    \mathbb{P}\left\{\sup_{1\leq l\leq N_E}\sup_{T_l\leq k\leq 2T_l-2}\left\|E_{k, p} \bar{Z}_{k, p}^{{\tr}} \bar{V}_{k, p}^{-\frac{1}{2}}\right\|_{2}^{2} \leq \operatorname{poly}\left(\bar{\sigma}_{\bar{R}},m,\beta,\log\left(\frac{1}{\delta_1}\right)\right) \log N\right\}\ge 1-\frac{\pi^2\delta_1}{6}.$$
    \item Conditioned on $\mathcal{E}_{PE}$ and $\mathcal{E}_e$, we have
    $$
    \mathbb{P}\left\{\sup_{1\leq l\leq N_E}\sup_{T_l\leq k\leq 2T_l-2}\left\|\bar{V}_{k,p}^{-\frac{1}{2}}\bar{V}_{k+1,p}^{\frac{1}{2}}\right\|_2^2\leq\operatorname{poly}\left(\frac{d}{M\sigma_R},\left\|a\right\|_2,\beta,\operatorname{log}\frac{1}{\delta_1},\bar{\sigma}_{\bar{R}}\right)\right\}\ge 1-\left(1+\frac{\pi^2}{6}\right)\delta_1.
    $$
    \item 
    Conditioned on $\mathcal{E}_{PE}$ and $\mathcal{E}_Z$, we have
    $$\mathbb{P}\left\{\sum_{k=T_{\text{init}}}^{N}\left\|\bar{V}_{k+1, p}^{-\frac{1}{2}} Z_{k+1,p}\right\|_{2}^{2}\leq \operatorname{poly}\left(M,m,\beta,\log\frac{1}{\delta}\right)\log^2 N\right\}\ge 1-\left(1+\frac{\pi^2}{6}\right)\delta_1.
    $$
\end{enumerate}
Together with the cross-term bound, we have
for probability at least $1-\left(3+\frac{5\pi^2}{6}\right)\delta_1$, there is
$$
\mathcal{R}_N\leq \operatorname{poly}\left(M,m,\beta,\log\frac{1}{\delta_1},d,\left\|a\right\|_2\right)\log^3(N).$$ Let $\delta = \left(3+\frac{5\pi^2}{6}\right)\delta_1$ then the result follows.

\input{Section-IV-v1}

\section{Additional Experiments}\label{secAdditionalExperiments}
In the additional experiments, without special statement, the system model is chosen to be the same as that in \cref{secSimulation}. 

\subsection{Comparison with low rank approximation based method}
\begin{figure}[h!]
    \centering
    \includegraphics[height=0.30\textwidth]{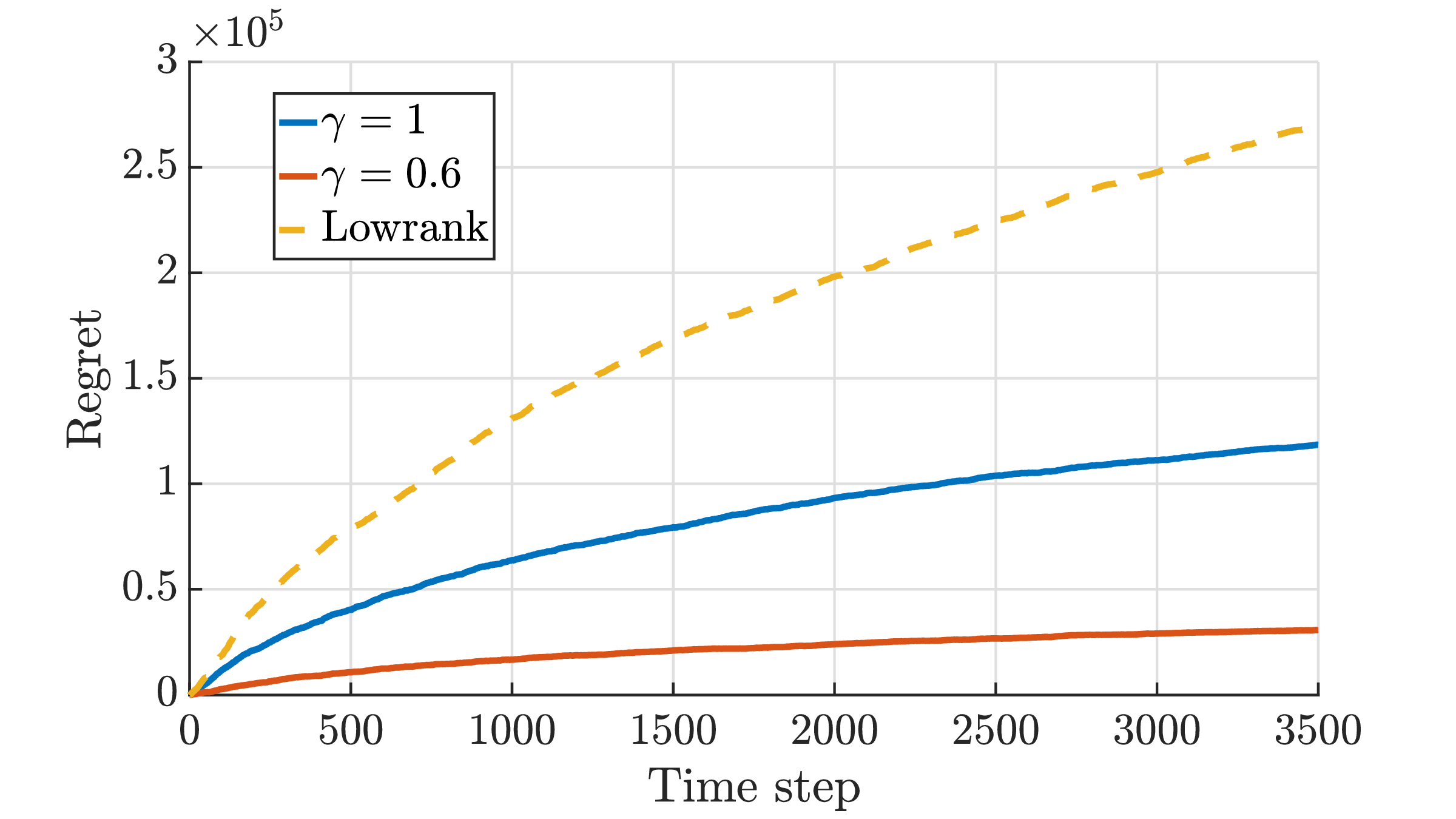}
    \caption{Illustration figure for the comparison of regret for different methods.}
    \label{figCompareLow}
\end{figure}

In the first supplementary experiment, we provide a fair comparison with the low rank approximation-based algorithm proposed in \cite{rashidinejad2020slip} by considering an ill-conditioned system, where low rank approximation techniques are used for dealing with the case when $\rho(A-LC)$ is close to 1. We apply the system parameters to be in severe condition, i.e.,
$$
A = \begin{bmatrix}
    0.98&0.8&0\\0&0.98&0.8\\0&0&0.9
\end{bmatrix},C = I_3,Q=I_3,R=100I_3.
$$
Then we can compute $\rho(A-LC)=0.78$. Note that the algorithm proposed in \cite{rashidinejad2020slip} requires a relatively long backward horizon length $p$ to guarantee the performance; we choose $\beta=6$, $T_{\text{init}}=500$, and $N_E=3$. The result is shown in \Cref{figCompareLow}, where the blue line represents the regret of the truncated algorithm in \cite{tsiamis9894660}, the yellow line represents the regret of the prediction algorithm in \cite{rashidinejad2020slip} and the red line represents the regret of the prediction algorithm proposed in this paper. From the comparison result we can see that our algorithm performs better than the low rank approximation based method even for noisy data and slow converge of $C(A-LC)^pL$, and we also verify that the performance of \Cref{algPrediction} is still better than that of \cite{tsiamis9894660} even when $\gamma<\rho(A-LC)$, the full potential of the forgetting based \Cref{algPrediction} still requires further discussion.

\subsection{Discussion on numerical stability}

In this subsection, we discuss the numerical robustness of the initialization process for each epoch.
Note that at the beginning of each time epoch with initial time step as $T_l=2^{l-1}T_{\text{init}+1}$, we need to perform the initialization process, i.e.,
\begin{equation}\label{directCalculation}
    \begin{aligned}
    \tilde{V}_{T_l-1,p}=&\lambda I+\sum_{t=p}^{T_l-1} D_pZ_{t, p} Z_{t, p}^{\tr}D_p,\\
    \tilde{G}_{T_l-1,p}=&\left(\sum_{t=p}^{T_l-1} y_{t} Z_{t, p}^{\tr}D_p\right) \tilde{V}_{T_l-1,p}^{-1},
\end{aligned}
\end{equation}
for the following online regression and prediction.
However, with the increase of time $T_l$, the cross term $\sum_{t=p}^{T_l-1} y_{t} Z_{t, p}^{\tr}D_p$ will be large, then the numerical error of taking matrix inverse $\tilde{V}_{T_l-1,p}^{-1}$, i.e., $\Delta \tilde{V}$ will have a significant effect on the parameter $\tilde{G}_{T_l-1,p}$. That is $$\Delta G=\sum_{t=p}^{T_l-1} y_{t} Z_{t, p}^{\tr}D_p\Delta \tilde{V},$$ which will have a significant effect on the regret $\mathcal{R}_N$, especially for the case with $\gamma=1$. 

To solve this numerical problem, we need to apply the iterative update law for computing $\tilde{G}_{T_l-1,p}$, that is
\begin{equation}\label{iterativeCalculation}
    \begin{aligned} \tilde{V}_{k,p}=&\tilde{V}_{k-1,p}+ D_pZ_{k, p} Z_{k, p}^{\tr}D_p,\\
    \tilde{\Tilde{y}}_k=&
\tilde{G}_{k-1,p} D_pZ_{k, p},\\ 
\tilde{G}_{k,p}=&\tilde{G}_{k-1,p}+\left(y_{k}-\tilde{y}_{k}\right) Z_{k, p}^{\tr}D_p \tilde{V}_{k,p}^{-1}.
\end{aligned}
\end{equation}
Here we use $\tilde{\tilde{y}}_k$  to indicate that the prediction is nominal and will only be used for initialization. 

\begin{figure}[h!]
    \centering
    \setlength{\abovecaptionskip}{8pt}
    \includegraphics[width=\textwidth]{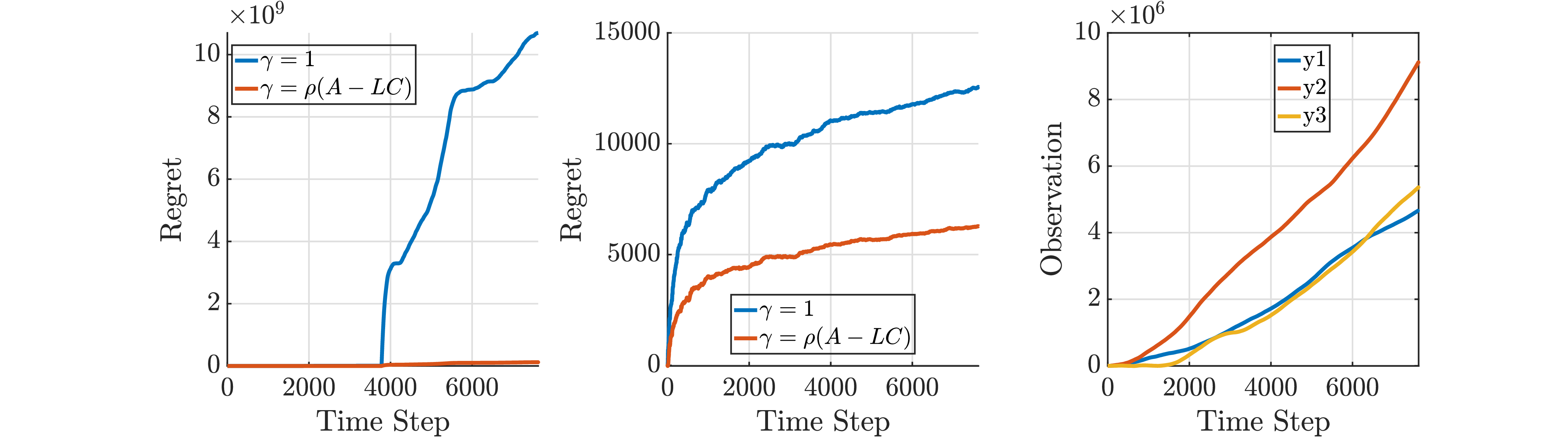}
    \caption{Illustration figure for the problem of numerical stability}
    \label{numerical}
    \vspace{-2mm}
\end{figure}

\cref{numerical} compare the performance of  \Cref{algPrediction} with different initialize law for $\tilde{G}_{T_l-1,p}$. The first sub-figure depicts the regret of \Cref{algPrediction} with direct update law \eqref{directCalculation}.
The second sub-figure depicts the regret of \Cref{algPrediction} with iterative update law \eqref{iterativeCalculation} and the last sub-figure is the trajectory of observation. From \cref{numerical}, we can see that the direct calculation of $\tilde{G}_{T_l-1,p}$ with \eqref{directCalculation} will cause a sharp increase in the regret, i.e., a large prediction error at the beginning of each epoch. Hence, from the perspective of numerical stability, the iterative update law \eqref{iterativeCalculation} performs much better than the direct update law \eqref{directCalculation}.

\subsection{Discussion on the selection of $\beta$}
In \Cref{sectionBeta}, we show that if the term $\beta$ is chosen as $\frac{2\kappa+1}{\log(1/\rho(A-LC))}$, then the bias term $C(A-LC)^p b_{k,p}$ will be sufficiently small such that the error induced by bias will also be neglectable.
Then the logarithmic regret of the online prediction \Cref{algPrediction} can be guaranteed. 

\begin{figure}[t]
    \centering
    \subfigure[Bias Cancellation]{\includegraphics[height = 0.26\textwidth]{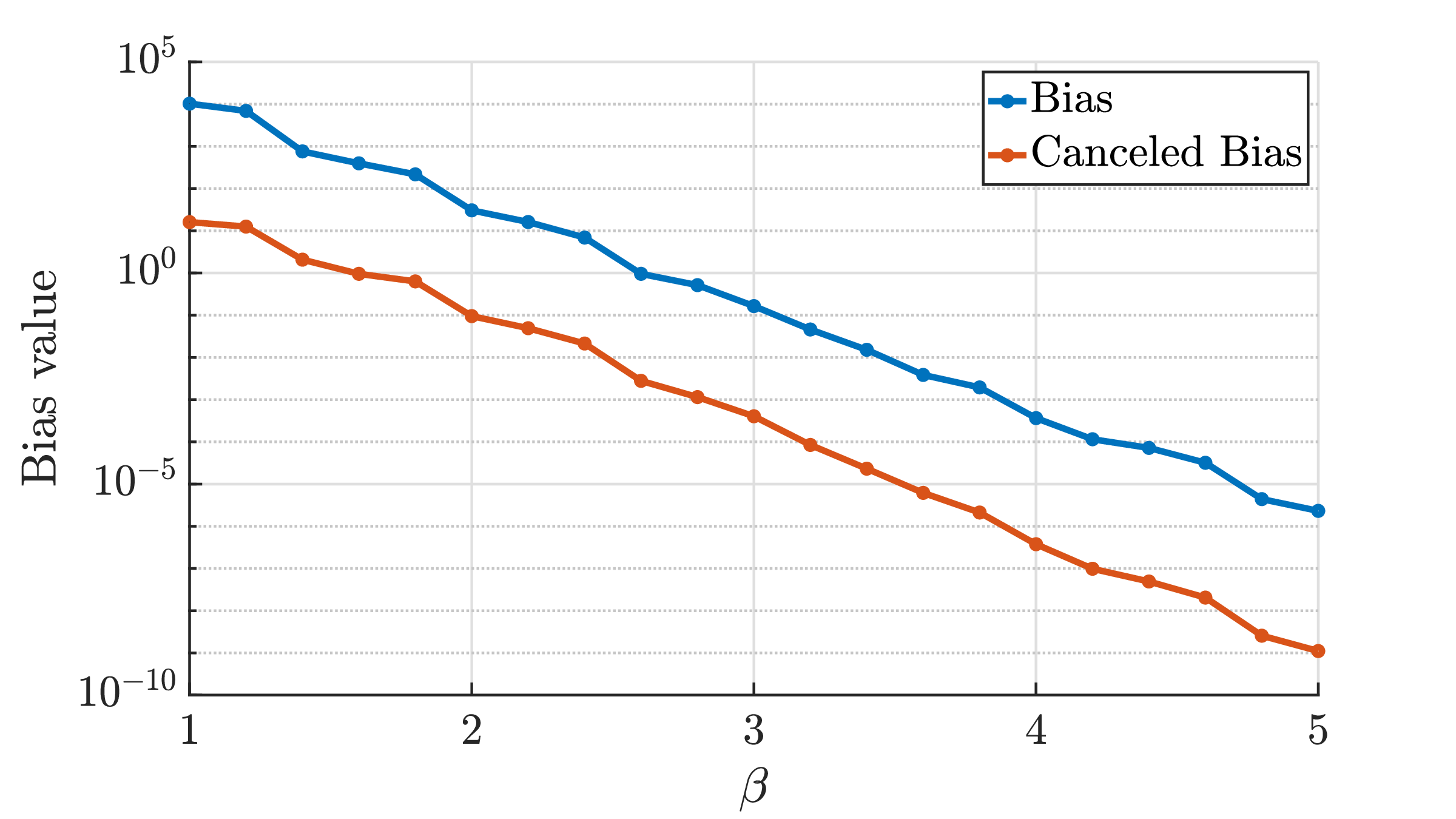} 
    \label{biasCancellation}} \hspace{3mm} \subfigure[Traditional Forget]{\includegraphics[height=0.26\textwidth]{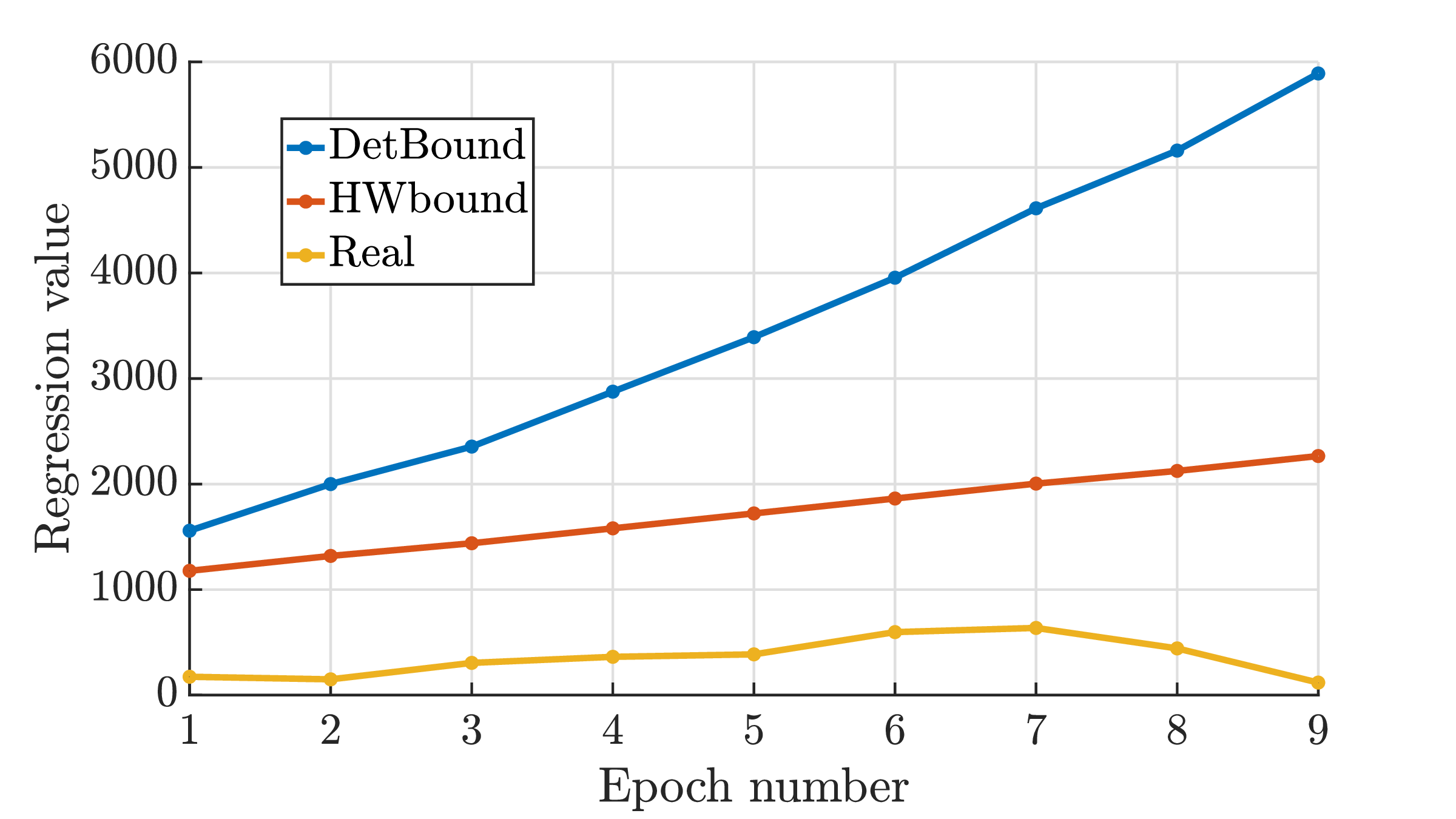}
    \label{figregressionBound}}
    \caption{Supplementary experiments}
    
\end{figure}

We find in numerical simulation that the bias error term $\sum_{l=p}^{k} b_{l, p} Z_{l, p}^{\tr} \bar{V}_{k, p}^{-1} Z_{k+1, p}-b_{k+1, p}$ has an effect of cancellation. That is, even if the term $b_{k+1, p}$ is large, the difference between $b_{k+1, p}$ and $\sum_{l=p}^{k} b_{l, p} Z_{l, p}^{\tr} \bar{V}_{k, p}^{-1} Z_{k+1, p}$ is much smaller than that of $b_{k+1, p}$; see the \cref{biasCancellation} for more details. 

In \cref{biasCancellation}, the blue line represents the average of the bias term $b_{k+1, p}$ over all $k$ in the last epoch with difference parameter $\beta$. While the red line represents the average of canceled bias term $\sum_{l=p}^{k} b_{l, p} Z_{l, p}^{\tr} \bar{V}_{k, p}^{-1} Z_{k+1, p}-b_{k+1, p}$ over $k$ with different $\beta$. As we can see from \Cref{biasCancellation} that the canceled bias term is much smaller than the real bias term $b_{k+1, p}$.  This phenomenon indicates that it is possible for us to choose a relatively smaller $\beta$ to reduce the length of the past horizon $p$ and balance a better trade-off between prediction error,  effect of overfitting and computation cost.

\subsection{Further experiments for theoretical verification}
In this subsection, we provide numerical verification for our theoretical statements.
In Fig.~\ref{figregressionBound}, we compare the upper bound of regression term $\big\|E_{k,p}\bar{Z}_{k,p}\bar{V}_{k,p}^{-\frac{1}{2}} \big\|_2^2$ provided in \cite{abbasi2011improved, tsiamis9894660, rashidinejad2020slip} and in this paper. 

In \cite{abbasi2011improved}, it is shown that the regression term $\big\|E_{k,p}\bar{Z}_{k,p}\bar{V}_{k,p}^{-\frac{1}{2}} \big\|_2^2$ is upper bounded by the determinant of $\bar{V}_{k,p}$, i.e.,
$$
\big\|E_{k,p}\bar{Z}_{k,p}\bar{V}_{k,p}^{-\frac{1}{2}} \big\|_2^2\leq O\Big(\log\frac{\det \bar{V}_{k,p}}{\delta\det \lambda D_p^{-2}}\Big)
$$
with high probability $1-\delta$. However, in \Cref{secMainProof}, we managed to prove that the dominant term of $\big\|E_{k,p}\bar{Z}_{k,p}\bar{V}_{k,p}^{-\frac{1}{2}} \big\|_2^2$ is around $\text{tr}\big(\bar{Z}_{k,p}\bar{V}_{k,p}^{-1}\bar{Z}_{k,p}^{\tr}\big)$. In \Cref{figregressionBound}, we compare the two different theoretical bounds with the real $\big\|E_{k,p}\bar{Z}_{k,p}\bar{V}_{k,p}^{-\frac{1}{2}} \big\|_2^2$, we select the value at $k=2T_l-2$, i.e., the end of each epoch, for comparison. Then we can see that the determinant bound provided by \cite{abbasi2011improved, tsiamis9894660, rashidinejad2020slip} is, to some extent, conservative. The bound provided in this paper is sharper, but there is still some potential to be optimized.   

\begin{figure}[t]
    \centering
    \includegraphics[width=0.98\textwidth]{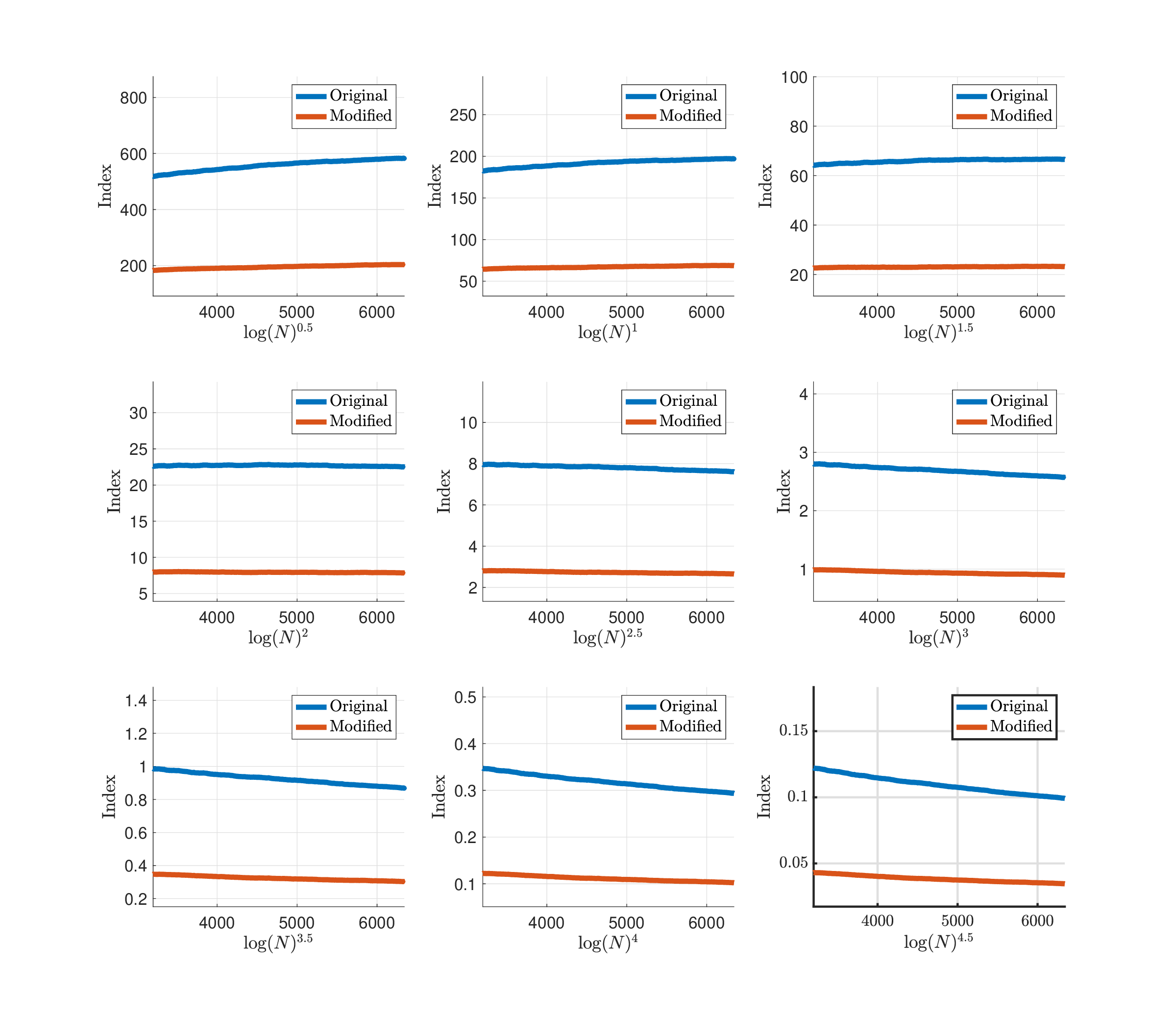}
    \caption{Illustration figure for the verification of theoretical regret bound.}
    \label{figVerifyRegret}
\end{figure}

Then we want to verify the theoretical result in Theorem \ref{thm1}. In \Cref{figVerifyRegret}, we compare the regret with different orders of $\log N$. The performance index is defined as 
$$
p(i,N) = \frac{\mathcal{R}_N}{\log^{i} N},
$$
where the blue line denotes the performance of the online prediction algorithm proposed in \cite{tsiamis9894660} and the red line denotes the performance of \Cref{algPrediction} proposed in this paper. It can be verified from this simulation that the regret of the online prediction algorithm appears to be $O(\log^2 N)$. The utilization of forgetting factor can improve the prediction performance, but cannot reduce the order of regret with $\log N$.

\section{Further Discussions}\label{secFurther}
\subsection{Extension to systems with control input}\label{secControl}
In this subsection, we provide a summary on how to directly extend the theoretical results and techniques proposed in this paper to the case with control input, i.e., to predict the stochastic system
    $$
    \begin{aligned}
x_{k+1}&=Ax_k+Bu_k + w_k, \quad  w_k \sim \mathcal{N}(0,Q)\\
y_{k}&=Cx_k+Du_k+ v_{k}, \quad v_k \sim \mathcal{N}(0,R).
\end{aligned}
    $$ 
    With a similar technique in \cref{subsecLog}, we can obtain the regression model as
    $$
    y_k=G_pZ_{k,p}+Du_{k}+C(A-LC)\hat{x}_{k-p}+e_k
    $$
    where $G$ and $Z_{k,p}$ are modified to be $G=[G^{(1)}_p,G^{(2)}_p]$, $Z_{k,p}=[Z_{k,p}^{(1)\tr},Z_{k,p}^{(2)\tr}]^\tr$ where
    $$
    \begin{aligned}
G^{(1)}_{p} &\triangleq\begin{bmatrix}
C(A-L C)^{p-1} L & \cdots & C L
\end{bmatrix}, \\
G^{(2)}_{p} &\triangleq\begin{bmatrix}
C(A-L C)^{p-1} (B-LD) & \cdots & C (B-LD)
\end{bmatrix},
    \end{aligned}
$$
and
$$
Z^{(1)}_{k,p}=\begin{bmatrix}
    y_{k-p}^\tr,\dots,y_{k-1}^\tr
\end{bmatrix}^\tr,\quad Z^{(2)}_{k,p}=\begin{bmatrix}
    u_{k-p}^\tr,\dots,u_{k-1}^\tr
\end{bmatrix}^\tr.$$ 

The forgetting process can be modified as
$$
\tilde{Z}_{k,p}=\begin{bmatrix}
    D_p^{(1)}&\\&D_{p}^{(2)}
\end{bmatrix}\begin{bmatrix}
    Z_{k,p}^{(1)}\\Z_{k,p}^{(2)}
\end{bmatrix},
$$
where $D_p^{(1)}$ and $D_p^{(2)}$ are diagonal matrices with proper dimension.
Then for uniformly bounded input $u_{k}$, we can verify that our algorithm directly guarantees a logarithmic regret $\mathcal{R}_N$ to the optimal predictor.
The regression process can directly follow the steps in \Cref{algPrediction}.

    \subsection{Matrix forgetting factor}
    For technical improvement, we can also consider applying the matrix-based forgetting factor to balance the regression model, i.e.,
    $$
    D_p=\operatorname{diag}(\Gamma^{p-1},\Gamma^{p-2},\dots,\Gamma,I),$$
    and the corresponding balanced model takes the form as
    \begin{align*}
\tilde{G}_p\triangleq \begin{bmatrix}
C(A-L C)^{p-1} L\Gamma^{-(p-1)}, \cdots, C L
\end{bmatrix}, 
\end{align*}
where the forgetting matrix $\Gamma$ can be designed with some system identification based techniques \cite{ljung1998system, tsiamis2019finite, zheng2020non}. Developing the prediction technique by potentially bridging the direct and indirect methods together to improve performance might be possible.

\end{document}

%% file: Preamb.tex
\usepackage{xcolor}
\usepackage{fullpage}
\usepackage{amsmath,amsthm,amssymb,amsfonts}
\usepackage{algorithm} %2e %algorithmic
\usepackage{algorithmic}
\usepackage{comment}
\usepackage[hidelinks]{hyperref}
\hypersetup{
    colorlinks=true,
    linkcolor=blue,
    filecolor=magenta,      
    urlcolor=cyan,
    pdftitle={Overleaf Example},
    pdfpagemode=FullScreen,
}

\usepackage[noblocks]{authblk}
\usepackage{tabularx}%table
\usepackage{booktabs}%table

\usepackage{mathtools}
\usepackage{tcolorbox}
\usepackage{bbm}
\usepackage{todonotes}
\usepackage{tensor}
\usepackage[font=small,labelfont=bf]{caption}
\usepackage{subcaption}
\usepackage{graphicx}
\usepackage{sectsty}

\usepackage{adjustbox} %table
\usepackage{enumitem} %indent
\usepackage{threeparttable}

\newcommand{\tr}{\mathsf{ T}}

\newtheorem{theorem}{Theorem}
\newtheorem{remark}{Remark}%[section]
\newtheorem{lemma}{Lemma}%[section]

\newtheorem{assumption}{Assumption}%[section]

\newcommand{\bsf}[1]{\textsf{\LARGE\textbf{#1}}}
\newcommand{\bigline}{\\ \vrule height2pt width 5 in depth 0pt\newline\noindent}

\definecolor{moccasin}{rgb}{0.98, 0.92, 0.84}
\newtcolorbox{mybox}{colback=moccasin,
colframe=moccasin}

\usepackage[nameinlink]{cleveref}
\crefname{equation}{}{}
\crefname{theorem}{Theorem}{Theorems}
\crefname{corollary}{Corollary}{Corollaries}
\crefname{example}{Example}{Examples}
\crefname{assumption}{Assumption}{Assumptions}
\crefname{lemma}{Lemma}{Lemmas}
\crefname{proposition}{Proposition}{Propositions}
\crefname{figure}{Figure}{Figures}
\crefname{table}{Table}{Tables}
\crefname{section}{Section}{Sections}
\crefname{appendix}{Appendix}{Appendices}
\Crefname{equation}{}{}
\Crefname{theorem}{Theorem}{Theorems}
\Crefname{corollary}{Corollary}{Corollaries}
\Crefname{example}{Example}{Examples}
\Crefname{lemma}{Lemma}{Lemma}
\Crefname{proposition}{Proposition}{Propositions}
\Crefname{figure}{Figure}{Figures}
\Crefname{table}{Table}{Tables}
\Crefname{section}{Section}{Sections}
\Crefname{appendix}{Appendix}{Appendices}

%% file: Section-I-Intro-v1.tex
\section{Introduction}
\label{submission}

The problem of sequential online prediction in linear dynamical systems is fundamental across many fields and has a long-standing history \cite{kalmanfilter}. Its applications span from control systems \cite{Andersonoptimal}, robotics \cite{barfoot2024state}, natural language processing \cite{belanger2015linear}, and computer vision \cite{coskun2017long}.  
The celebrated Kalman filter \cite{kalmanfilter} predicts the future system state and observation by properly filtering the past observations. When the underlying dynamics are linear with Gaussian noise, the Kalman filter achieves optimal prediction among all possible filters by minimizing the mean square error. Many structural properties of the Kalman filter, including stability and statistical properties, have been well-established \cite{Andersonoptimal}. Extensions to nonlinear systems have also been well developed in the past \cite{ ribeiro2004kalman, arasaratnam2009cubature}. 

All these classical developments on the Kalman filter and its variants rely on exact knowledge of the system model and noise statistics. However, when the system model and/or noise statistics are \textit{unknown}, learning-based approaches are required to predict future observations. In this paper, we consider the problem of predicting observations in an \textit{unknown}, \textit{partially observed}, \textit{nonexplosive} linear dynamical system driven by Gaussian noise using \textit{finite} past observations.  
Learning for prediction is not a new topic and has a rich list of literature; see \cite{ljung1998system,sayed2003fundamentals} for classical textbooks. This topic has recently received renewed interest from the online learning perspective \cite{hazan2017learning,ghai2020no,tsiamis9894660}. 

The methodology of learning for prediction can be broadly divided into two categories: the \textit{indirect} method and the \textit{direct} method. The indirect method first exploits the past data to build an approximation of the nominal system \cite{tsiamis2019finite, zheng2020non, lee2022improved} and then designs a filter for prediction based on the identified model. However, learning directly the model parameters often leads to a nonlinear and nonconvex problem even for linear dynamical systems \cite{hardt2018gradient}. Many adaptive filtering techniques have been developed to address unknown models or changing environments \cite{lai1991recursive,sayed2003fundamentals}. These developments are typically based on extended least squares and variants for parameter estimation and state prediction, and the classical analysis is often based on {asymptotic} tools assuming {infinite} samples.     

The direct method aims to design a filter by directly learning the optimal prediction policy from data. This perspective has gained increasing attention in the online learning framework \cite{anava2013online, hazan2017learning, kozdoba2019line, rashidinejad2020slip}. In this context, the goal of model estimation is replaced by \textit{regret} minimization in online prediction, which measures the loss compared to the best prediction in hindsight. 
For example, 
a low-rank approximation technique was proposed in \cite{hazan2017learning},
and \cite{kozdoba2019line} utilized structural properties of the Kalman filter to derive a regression model where the output is the optimal prediction and the input is previous observations. The regret guarantees in \cite{hazan2017learning,kozdoba2019line} are sublinear. A recent study \cite{li2024regret} also presents an online learning strategy to achieve a sublinear regret. The first algorithmic regret was established in \cite{tsiamis9894660}, where a truncated regression model was used. Another study \cite{rashidinejad2020slip} also established a logarithmic regret for a low-rank spectral approximation strategy; see \Cref{sample-table} for a comparison.

\begin{table}[t]
\caption{Regrets in Recent Online Prediction Algorithms} \vspace{-3mm}
\label{sample-table}
\begin{center}
\begin{small}
\begin{sc}
\begin{tabular}{p{6cm}p{3.0cm}p{2.0cm}p{2.2cm}}
\toprule
\textnormal{Algorithms}  & \textnormal{Regret}  & \textnormal{Gaussian}& \textnormal{Forgetting}\\
\midrule
\textnormal{Hazan et al.}, 2017 \cite{hazan2017learning}& $\tilde{O}(\sqrt{N})$ & $\times$& $\times$ \\
\textnormal{Kozdoba et al.}, 2019 \cite{kozdoba2019line}& $\tilde{O}(\sqrt{N}+\epsilon N)$&$\times$&$\times$\\
\textnormal{Li et al.}, 2024 \cite{li2024regret}&  $\tilde{O}(\sqrt{N})$ & $\times$& $\times$\\

\textnormal{Rashidinejad et al.}, 2020 \cite{rashidinejad2020slip}& $O(\log^{11} N)$& $\surd$& $\times$ \\

\textnormal{Tsiamis and Pappa}, 2023 \cite{tsiamis9894660} & $O(\log^6 N)$& $\surd$& $\times$\\
\textnormal{Our paper} & $O(\log^3 N)$& $\surd$& $\surd$ \\
\bottomrule

\end{tabular}
\end{sc}
\end{small}
{\footnotesize 
\begin{tablenotes}
            In this table, ``Gaussian'' means whether the system noises are assumed to be Gaussian; ``Forgetting'' means whether any forgetting strategy is used. 
\end{tablenotes}}
\end{center}
\vspace{-5mm}
\end{table}

\begin{figure}[t]
    \centering
    \setlength{\abovecaptionskip}{2pt}
    \includegraphics[height=0.26\textwidth]{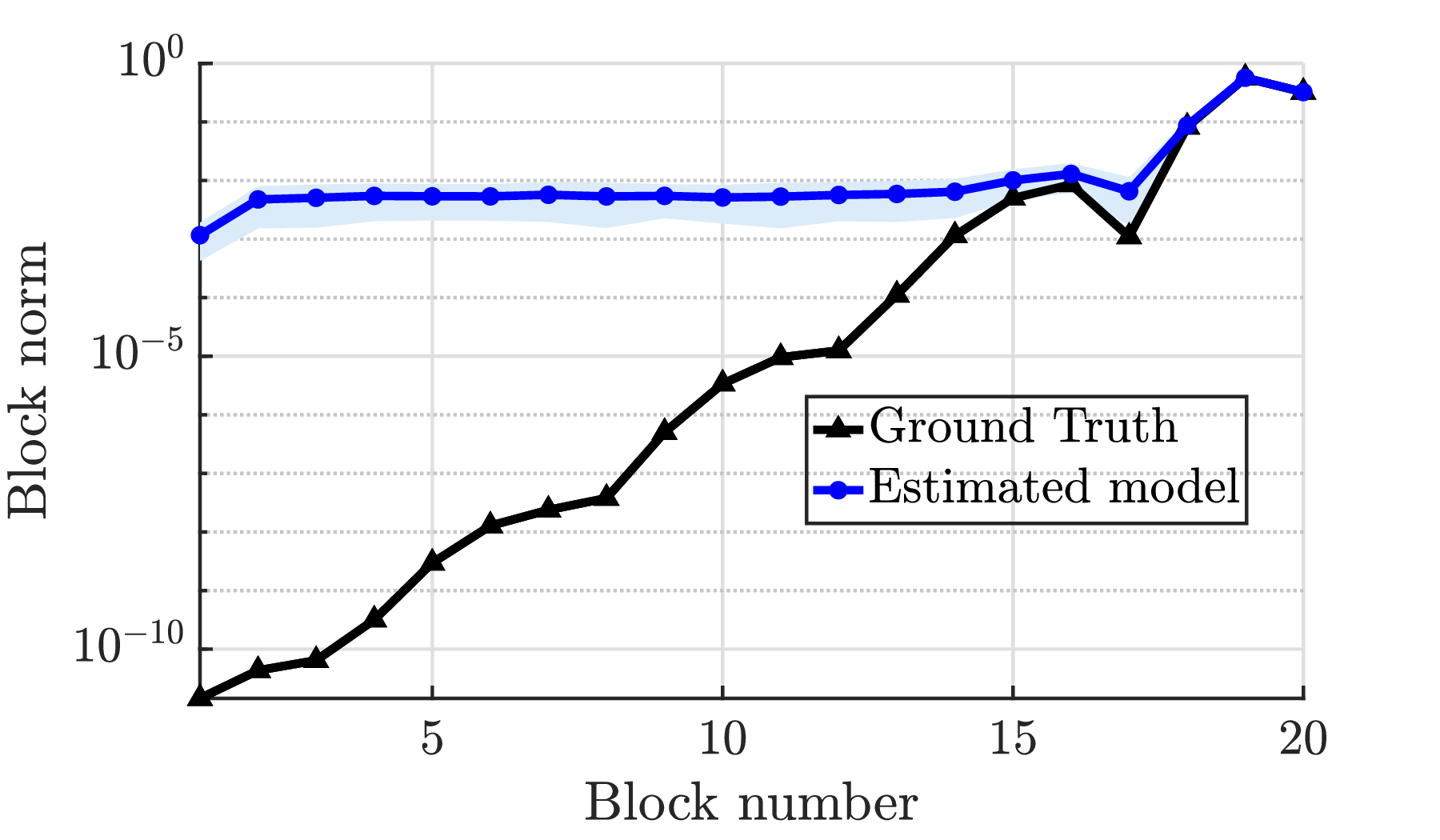}
    \caption{Illustration of the overfitting effect in a simple example. The black curve shows the ground truth that has a highly imbalanced structure, and the blue curve is the estimated model using method provided in \cite{tsiamis9894660} that overfits the small blocks. }
    
    \vspace{-2mm}
    \label{figMotivation}
\end{figure}

One main challenge in online prediction is handling non-explosive systems, where the system state exhibits a \textit{long-term memory} effect and may even grow at a polynomial rate. This long-term memory effect can introduce a persistent bias in the regression model \cite{kozdoba2019line}. To mitigate the bias, it is crucial to gradually expand the past data horizon for prediction \cite{tsiamis9894660}. However, due to the recursive nature of the Kalman filter, the resulting regression model becomes highly \textit{imbalanced}, with each block potentially differing by several orders of magnitude. This imbalance can easily lead to \textit{overfitting} (see \Cref{figMotivation}), thus degrading the regret performance of the existing online prediction algorithms \cite{hazan2017learning, kozdoba2019line,rashidinejad2020slip,tsiamis9894660}.         
Forgetting strategies, also known as discounting, are widely used to mitigate overfitting and adapt to non-stationary environments \cite{lee2019concurrent,zinkevich2003online,jacobsen2024online}. 
These~techniques prioritize recent information over outdated data, ensuring models remain responsive to changing dynamics. A common approach is exponential forgetting, where past observations are down-weighted using a forgetting factor, preventing the dominance of earlier data in recursive updates. This technique is frequently employed in recursive least squares and adaptive Kalman filtering \cite{johnstone1982exponential,sayed2003fundamentals,ciochina5206117forgetting,lai2024generalized}. However, most existing forgetting strategies focus on re-weighting past and recent data, which does not effectively address the {imbalanced structure} inherent in online learning of the Kalman filter. 

In this paper, we address the imbalanced regression model in model-free online learning of the Kalman filter. 
We also aim to derive a sharper logarithmic regret bound of our online prediction algorithm. 
Our technical contributions are twofold. 
\vspace{-1mm}
 \begin{itemize}
     \item 
     \setlength{\itemsep}{0pt}
 At the algorithmic level, we introduce an \underline{O}nline \underline{P}rediction~with \underline{F}orgetting (\texttt{OPF}) algorithm which uses an {exponentially forgetting} strategy to properly re-balance the regression model. 
 Our forgetting approach aims to directly balance the {regression model}, whereas most existing discounting and forgetting methods primarily aim to re-weight the online data   \cite{jacobsen2024online,lai2024generalized}. 
We further show that our forgetting strategy is equivalent to a generalized Ridge regression, which introduces an {inductive bias} reflecting the exponentially decaying structure in the regression model. 
    Our \texttt{OPF} algorithm effectively mitigates the overfitting effect and demonstrates improved performance compared to existing online prediction algorithms \cite{tsiamis9894660, hazan2017learning, rashidinejad2020slip}.
  \item  At the level of theoretical guarantees, 
     the best theoretical regret bound previously established is $O(\log^6 N)$ for marginally stable system \cite{tsiamis9894660}, where $N$ is the number of observations. In this paper, we establish a much shaper regret bound of $O(\log^3 N)$ by new analysis techniques, including Hanson-Wright inequality based martingale analysis and self-cancellation property of accumulation error. \Cref{sample-table} lists a comparison of the bounds in recent online prediction algorithms. 
     Note that our use of the forgetting factor may not directly reduce the order of the logarithmic regret bound. Instead, it effectively controls accumulation error and achieves a better trade-off between regression error and regularization error. This significantly improves the constant factors in the regret bound, as demonstrated in our numerical experiments. 
 \end{itemize}

The rest of this paper is organized as follows. \Cref{section:background} presents preliminaries on the Kalman filter. We introduce the new \texttt{OPF} algorithm in \Cref{section:balancing}, and its regret guarantee is established in \Cref{Section:Regret-guarantee}. Numerical experiments are presented in \cref{secSimulation}. \Cref{section:conclusion} concludes this paper. Further discussions and proof details are presented in the appendix.  

%% file: Section-II-v1.tex
\section{Preliminaries and Problem Formulation}
\label{section:background}

\subsection{Background on the Kalman Filter}

{Consider} a partially observed linear stochastic system
\begin{equation}\label{linearSystem}
\begin{aligned}
x_{k+1}&=Ax_k + w_k, \quad  w_k \sim \mathcal{N}(0,Q)\\
y_{k}&=Cx_k + v_{k}, \quad v_k \sim \mathcal{N}(0,R), 
\end{aligned}
\end{equation}
where $x_k\in\mathbb{R}^n$ is the state vector,  and $y_{k}\in\mathbb{R}^{m}$ is the output measurement, $A \in \mathbb{R}^{n\times n}$ is the state transition matrix, and $C \in \mathbb{R}^{m \times n}$ is the observation matrix. The time series $w_k$ and $v_k$ denote the process and measurement noises, respectively. We assume they are zero-mean independent identically distributed (i.i.d) Gaussian variables with covariances $Q \succ 0$ and $R \succ 0$ respectively. 

An important problem is to predict the observation $y_{t+1}$ based on past measurements,  a process commonly referred to as \textit{filtering} or \textit{prediction}\footnote{Depending on the context, {filtering} and {prediction} may have slightly different meanings \cite{Andersonoptimal}; we here treat them as equivalent concepts. Also, all our discussions can be applied to linear systems with inputs; see \cref{secControl}. }. 
{To guarantee the existence of a stable filter}, 
we make a standard assumption.
\begin{assumption}\label{aspOb-main-text}
    The system pair $(A, C)$ is detectable.
\end{assumption}

Let  $\mathcal{F}_{k} \triangleq \sigma\left(y_{0}, \ldots, y_{k}\right)$  be the filtration generated by the observations  $y_{0}, \ldots, y_{k}$. Given past observations up to time $k$, we consider the problem of optimal prediction  $\hat{y}_{k+1}$ at time  $k+1$ in the minimum mean-square error sense:  
\begin{equation}\label{MMSEProblem}
    \hat{y}_{k+1} \triangleq \arg \min _{z \in \mathcal{F}_{k}} \mathbb{E}\left[\left\|y_{k+1}-z\right\|_{2}^{2} \mid \mathcal{F}_{k}\right].
\end{equation}
Thanks to the foundational work \cite{kalmanfilter}, 
it is now well-known that the optimal predictor takes a recursive form, known as the \textit{Kalman filter}, 
\begin{equation}\label{OptimalPDT-main}
\begin{aligned}
\hat{x}_{k+1} & =A \hat{x}_{k}+L_k\left( y_{k}-\hat{y}_{k}\right) 
,\;\;\hat{x}_{0}=0 \\
\hat{y}_{k} & =C \hat{x}_{k}, 
\end{aligned}
\end{equation}
where $L_k=AP_kC^\tr\left(CP_kC^\tr+R\right)^{-1}$ with $P_k$ satisfying 
\begin{equation}\label{RicRecursion-main}
    P_{k+1}=AP_k A^\tr+Q-AP_kC^\tr\left(CP_kC^\tr+R\right)^{-1}CP_kA^\tr. 
\end{equation}
With \Cref{aspOb-main-text},  $P_k$ in the recursion \eqref{RicRecursion-main} converges to a steady-state $P$ exponentially fast, which leads to the~well-studied  algebraic Riccati equation below  \cite{lancaster1995algebraic}
\begin{equation}\label{Ric}
    P=APA^\tr+Q-APC^\tr\left(CPC^\tr+R\right)^{-1}CPA^\tr.
\end{equation}
The steady-state Kalman gain is denoted as $L$. 
We remark that the error $e_{k+1} = y_{k+1} - \hat{y}_{k+1}$ is often referred to as the \textit{innovation} noise process, which represents the amount of new information $y_{k+1}$ brings beyond what is already known from previous observations up to $y_{k}$.  

\subsection{Model-free Prediction and Regret}\label{subsecLog}

The Kalman filter in \eqref{OptimalPDT-main} requires the system matrices $A$ and $C$ as well as noise statistics $Q$ and $R$. This is thus a \textit{model-based} prediction, with its performance heavily dependent on the availability and accuracy of the system model.  

In this paper, we focus on \textit{model-free} online learning algorithms to predict observations based only on past measurements, without requiring any knowledge of the system matrices or noise covariances. 
Following \cite{tsiamis9894660,rashidinejad2020slip}, we quantify the performance of the online prediction in terms of the \textit{regret} measured against the Kalman filter \eqref{OptimalPDT-main} that has full system knowledge.  
The goal is to design an online algorithm $\tilde{y}_k=f_k(y_0,\dots,y_{k-1})$ such that with probability at least $1-\delta$, the following regret
\begin{equation}\label{regret}
    \mathcal{R}_{N} \triangleq \sum_{k=1}^{N}\left\|y_{k}-\tilde{y}_{k}\right\|^{2}-\sum_{k=1}^{N}\left\|y_{k}-\hat{y}_{k}\right\|^{2},
\end{equation}
where $\hat{y}_{k}$ is the steady-state  Kalman filter's prediction, 
is bounded by $\operatorname{poly}(\log 1 / \delta) o(N)$. Note that, due to the exponential convergence of the Kalman filter to its steady-state, it is sufficient to compare the online algorithm with the steady-state predictor $\hat{y}_k$. 
The regret benchmark~in~\eqref{regret}~is~much stronger than that in \cite{hazan2017learning}. Still, recent advances in \cite{tsiamis9894660,rashidinejad2020slip} have shown that it is possible to obtain \textit{logarithmic} regrets. 

One key observation is that the Kalman filter \eqref{OptimalPDT-main} has a \textit{linear regression} structure after expanding the past $p$ observations: 
\begin{equation}\label{regression}
    y_{k}=G_{p} Z_{k, p}+C(A-L C)^{p} \hat{x}_{k-p}+e_{k},
\end{equation}
where $e_k = y_k - \hat{y}_k$, and $Z_{k, p}$ denotes the past $p$ observations at time $k$, i.e., 
$
Z_{k, p} \triangleq\begin{bmatrix}
    y_{k-p}^{\tr}\!\!\!\!\!& \ldots & \!\!\!y_{k-1}^{\tr}
\end{bmatrix}^\tr,
$
and $G_{p}$ denotes the Markov parameters of the closed-loop filter: 
\begin{equation} \label{eq:regressor}
G_{p} \triangleq\begin{bmatrix}
C(A-L C)^{p-1} L, & \cdots, & C L
\end{bmatrix} \in \mathbb{R}^{m \times pm}.
\end{equation}
To simplify discussions, we make another assumption.  
\begin{assumption}\label{aspDiagonal} (\cite{tsiamis9894660,rashidinejad2020slip})
    The matrix $A-LC$ is diagonalizable.
\end{assumption}

While \eqref{regression} is straightforward to derive based on the structure of Kalman filter \eqref{OptimalPDT-main}, this linear regression model provides sufficient insights: the system output at time $k$ can be viewed as a linear combination of the past observations with an additional bias term $C(A-L C)^{p} \hat{x}_{k-p}$ and the innovation noise process $e_k$. Under \Cref{aspOb-main-text}, it is well known that 1) the spectral radius satisfies $\rho(A-L C) < 1$, and 2) $e_k$ are mutually uncorrelated Gaussian random variables with zero-mean \cite{Andersonoptimal}. Thus, the model-free prediction can be reduced to an {online least-squares} problem  \eqref{regression}. Similar ideas can be traced back to subspace identifications  \cite{bauer1999consistency}.  

\subsection{Long-term Memory and Forgetting}
\label{secMotivation}

With \eqref{regression}, it is clear that the problem of model-free prediction for system \eqref{linearSystem} falls within the framework of {online least squares} \cite{draper1998applied}. Two recent works \cite{tsiamis9894660,ghai2020no} have used basic recursive auto-regression algorithms to estimate $G_{p}$ and make online predictions. To reduce the dimension of $G_{p}$,  \cite{rashidinejad2020slip} has embedded spectral approximation into algorithm design based on the concept of Kolmogorov width.  

However, the practical performance of these existing approaches can easily deteriorate due to the {highly imbalanced nature} of the regression model \eqref{regression}. Indeed, for nonexplosive systems $\rho(A) \leq 1$, the state $\hat{x}_{k-p}$ has a {long-term memory effect} and leads to a persistent bias~term. A suitable past horizon length $p$ is required to mitigate $\hat{x}_{k-p}$ in \eqref{regression}. Since $\rho(A - LC) < 1$, the magnitude of each block in $G_p$ exponentially decays to $0$, i.e.,  $\left\|C(A-LC)^t L\right\|_2\leq M\rho(A-LC)^t$ under \Cref{aspDiagonal}, where $M$ is a constant. 
This imbalance can easily lead to {overfitting} during online learning, thereby degrading the prediction accuracy of methods in \cite{kozdoba2019line,tsiamis9894660,rashidinejad2020slip,ghai2020no}. \Cref{figMotivation} illustrates this overfitting effect on a simple example. The magnitude of the true regression model $G_p$ (represented by the black line) is significantly smaller than the estimates $\tilde{G}_p$ from the algorithm in  \cite{tsiamis9894660}.
The basic least-squares algorithms in \cite{tsiamis9894660,rashidinejad2020slip,ghai2020no} can easily overfit the small values of blocks $C(A-LC)^t L$.   

In this paper, our key idea is to use an {exponentially forgetting} strategy to properly re-balance the regression model and mitigate overfitting, thus improving the online prediction performance. As we will show in \cref{section:balancing}, our forgetting approach aims to directly balance the {regression model} itself, whereas most existing discounting and forgetting methods primarily aim to re-weight the online data, as seen in the recent works  \cite{jacobsen2024online,lai2024generalized}. 
We further show that our forgetting strategy is equivalent to a generalized Ridge regression, which introduces an {inductive bias} reflecting the exponentially decaying structure \eqref{eq:regressor} in the regression \eqref{regression}.

%% file: Section-III-v1.tex
\section{Online Prediction with Forgetting} \label{section:balancing}

This section introduces a new \underline{O}nline \underline{P}rediction~with \underline{F}orgetting (\texttt{OPF}) algorithm. We also show that our forgetting strategy is equivalent to a generalized ridge regression with an inductive bias to re-balance the regression model. 

\subsection{Online Balanced Regression via Forgetting} 
\label{designIdea}

Our approach begins by addressing the exponentially decaying structure of the regressor $G_p$ in \eqref{eq:regressor}.
The key idea is to appropriately rescale the blocks as follows  
\begin{align*}
\tilde{G}_p\triangleq \begin{bmatrix}
\frac{1}{\gamma^{p-1}}C(A-L C)^{p-1} L,\; \cdots,\; \frac{1}{\gamma} C(A-L C) L,\;\, C L
\end{bmatrix}, 
\end{align*}
where $\gamma \in (0,1)$. If $\gamma$ is close to $\rho(A-L C)$, then the magnitude of each bock $\frac{1}{\gamma^t}C(A-L C)^{t} L$ will remain relatively \textit{balanced} across all terms. 

With this idea, we rewrite the regression model \eqref{regression} as 
\begin{equation}\label{regression-balanced}
    y_{k}=\underbrace{\tilde{G}_{p} \tilde{Z}_{k, p}}_{\text{Balanced regression}}\!\!+\;\;\underbrace{C(A-L C)^{p} \hat{x}_{k-p}}_{\text{Long-term bias}}\;\;+\underbrace{e_{k}}_{\text{Innovation noise}},
\end{equation}
where we have $\tilde{G}_{p} = G_pD_p^{-1}$ and $ \tilde{Z}_{k, p} := D_p  {Z}_{k, p}$ with 
\begin{equation}\label{equationDp}
D_p=\text{diag}(\gamma^{p-1},\gamma^{p-2},\dots,\gamma,1)\otimes I_m
\end{equation}
being a re-weighting matrix. The new regression model \eqref{regression-balanced} does not change the {long-term bias term} $C(A-L C)^{p} \hat{x}_{k-p}$ or the {innovation noise term} $e_k$. 
It is also worthwhile noting that the balanced regression \eqref{regression-balanced} aims to express the output at time $k$ as a linear combination of the past observations with {exponentially forgetting}. In other words, we have 
\begin{equation} \label{eq:samples-after-scaling}
    \tilde{Z}_{k,p} = D_p Z_{k,p}= \begin{bmatrix}
    \gamma^{p-1}y_{k-p}\\\gamma^{p-2}y_{k-p+1}\\\vdots\\y_{k-1}
\end{bmatrix}. 
\end{equation}
Instead of optimizing over the original Markov parameters $G_p$ in \eqref{eq:regressor} which easily leads to overfitting, we optimize over the re-balanced $\tilde{G}_p$ in \eqref{regression-balanced}. Note that optimizing over $\tilde{G}_p$ is known as \textit{improper learning} in the context of online learning \cite{foster2018logistic}, as we do not directly learn the true model parameters in the hypothesis class  $\mathcal{H}=(A, C, Q, R)$ but re-parameterize and learn over a different class $\tilde{\mathcal{H}}$. 

We learn a least-squares estimate $\tilde{G}_{k,p}$ by regressing~$y_t$ to past outputs $\tilde{Z}_{t,p}, t \leq k$  with {exponentially forgetting},~i.e., 
\begin{equation} \label{eq:regression-update}
   { \tilde{G}_{k,p} = \sum_{t = p}^k y_t \tilde{Z}_{t,p}^\tr \left(\lambda I + \sum_{t=p}^k \tilde{Z}_{t,p}\tilde{Z}_{t,p}^\tr\right )^{-1},} 
\end{equation}
where $\lambda > 0$ is a regularization parameter and the matrix 
\begin{equation}\label{eq:gramMatrix}
    \tilde{V}_{k,p}\triangleq\lambda I + \sum_{t=p}^k \tilde{Z}_{t,p}\tilde{Z}_{t,p}^\tr
\end{equation}
is called a Gram matrix, which contains the collected information of all the past knowledge $\tilde{Z}_{k,p}$.
 We then predict the next observation by computing 
\begin{equation} \label{eq:OPF-prediction}
    \tilde{y}_{k+1} = \tilde{G}_{k,p}\tilde{Z}_{{k+1},p}. 
\end{equation}

For nonexplosive systems $\rho(A) \leq 1$, the state $\hat{x}_{k-p}$ retains a {long-term memory effect}, and its value may even grow at a polynomial rate. A persistent bias error could result in linear regret. Fortunately, classical Kalman filter theory provides two key insights: 1) the innovation noise $e_k$ is zero mean and mutually uncorrelated, and 2) $\rho(A - LC) < 1$. The fact that the spectral radius is less than one enables us to manage the accumulation of long-term bias errors effectively. Specifically, we can gradually increase the past horizon $p$, an approach aligned with the strategy in \cite{tsiamis9894660} and also motivated by a common ``doubling trick'' \cite{cesa2006prediction}. In particular, 
the entire time horizon is partitioned into multiple epochs, with each successive epoch being twice as long as the previous one. Within each epoch, the value of $p$ past horizon remains constant. Given that $\rho(A - LC)^p$ is decreasing exponentially, it is sufficient to slowly increase the past horizon length as $p = \mathcal{O}(\log T)$, where $T$ is the length of each epoch. 

The pseudocode of our online prediction with forgetting (\texttt{OPF}) is listed in \Cref{algPrediction}, which has  two main phases: \textit{Warm-Up} and \textit{Online Prediction}. In the warm-up phase, we collect a trajectory of observations of length $T_{\mathrm{init}}$. In the phase of online prediction, we first initialize the parameter $p$, $D_p$, $\tilde{V}_{T_l-1,p}$ and $\tilde{G}_{T_l-1,p}$ at the beginning of each epoch $T_l$.  At each time step $k$, we update the prediction $\tilde{y}_k$  using \eqref{eq:OPF-prediction} and then observe the new observation ${y}_k$. 
Within each epoch, the predictor can be computed recursively
\begin{subequations} \label{eq:predictor-update}
\begin{align}
\tilde{V}_{k,p}&=\tilde{V}_{k-1,p}+D_pZ_{k, p} Z_{k, p}^{\tr}D_p, \\
\tilde{G}_{k,p}&=\tilde{G}_{k-1,p}+\left(y_{k}-\tilde{y}_{k}\right) Z_{k, p}^{\tr}D_p \tilde{V}_{k,p}^{-1}. 
\end{align}
\end{subequations}

\Cref{algPrediction} requires no knowledge of the system model~or noise covariances. Even the state dimension $n$ is not needed as we only predict the output. Note that the past~horizon $p=\beta \log T_l$ deals with the long-term memory $\tilde{x}_{k-p}$ in~\eqref{regression-balanced}. Selecting a larger $\beta$ leads to a smaller bias error but increases the estimation variance of $\tilde{G}_p$ due to larger freedom. 

\begin{remark}[Forgetting strategies] \label{rmkTraditional}
The idea of forgetting factor is widely used in online learning \cite{paleologu2008robust, jacobsen2024online, zhangdiscounted, tsiamis2024predictive}. To our knowledge, most existing studies only focus on balancing current and past data by applying a horizon-dependent forgetting factor \textit{uniformly} to each sample $\alpha^{k-t} Z_{t,p}$. In our context, this leads to  an updating law below:
\begin{equation}\label{traditionalForget}\tilde{G}_{k,p} \!=\! \sum_{t = p}^k \alpha^{k-t}y_t Z_{t,p}^\tr \left(\lambda I \!+\! \sum_{t=p}^k \alpha^{k-t}Z_{t,p}Z_{t,p}^\tr\right )^{-1}\!\!\!\!.
\end{equation}
Here, $\alpha \in (0,1)$ is used to distinguish it from $\gamma$.
However, \eqref{traditionalForget} can result in information loss due to exponential forgetting, which, in turn, increases prediction errors, particularly in marginally stable systems \citep[Lemma D.5]{tsiamis2024predictive}. In contrast, our \texttt{OPF} method leverages the forgetting factor to provide a \textit{non-uniform} scaling for each sample $\tilde{Z}_{k,p}=D_p Z_{k,p}$. This approach effectively balances the regression model with no information loss, thereby reducing the overfitting effect. Both theoretical guarantees in \Cref{Section:Regret-guarantee} and numerical experiments in \Cref{secSimulation} confirm our intuition.  \hfill $\square$
\end{remark}

\begin{algorithm}[tb]
   \caption{Online Prediction with Forgetting (\texttt{OPF})}\label{algPrediction}
\begin{algorithmic}[1]
   \STATE {\bfseries Input:} parameter $\beta, \lambda, \gamma, T_{\text {init }}, N_E$ 
   
   \COMMENT{\textit{Warm Up}:} 
   \FOR{$k=0$
   {\bfseries to }$T_{\text{init}}$}
   \STATE Observe $y_k$;
   \ENDFOR
   
   \COMMENT{\textit{Recursive Online Prediction:}}
    \FOR{$l=1$ {\bfseries to} $N_E$}
    \STATE
    Initialize 
      $T_l=2^{l-1} T_{\text {init }}+1,\;\; p=\beta \log T_l,$ $D_p$ \text{in \eqref{equationDp}}
    \STATE
    Re-balance  $\tilde{Z}_{t,p}=D_pZ_{k,p}$ in \eqref{eq:samples-after-scaling} for $p\leq t< T_l$; 
    \STATE
    Compute $\tilde{V}_{T_l-1},p$ in \eqref{eq:gramMatrix} and  $\tilde{G}_{T_l-1,p}$ in \eqref{eq:regression-update}; 
    \FOR{$k=T_l$ \textbf{to} $2 T_l-2$}
    \STATE Predict $\tilde{y}_{k}=\tilde{G}_{k-1,p} \tilde{Z}_{k,p}$; 
    \STATE
    Observe $y_{k}$;
    \STATE
    Update $\tilde{V}_{k,p}$ and $\tilde{G}_{k,p}$ recursively as \eqref{eq:predictor-update}. 
    \ENDFOR
    \ENDFOR   
\end{algorithmic}
\end{algorithm}

\subsection{Inductive Bias from Forgetting}

We here show that the exponentially forgetting strategy in our \texttt{OPF} algorithm is equivalent to solving
a generalized ridge regression with appropriate regularization. This regularization introduces an inductive bias {that captures our prior belief about the underlying imbalanced structure}, such that our \texttt{OPF} algorithm never forgets.  

Recall the re-scaled data samples at time step $k$ in \eqref{eq:samples-after-scaling}, i.e., $\tilde{Z}_{k,p}=D_{p} Z_{k, p}$. We define the two matrices as
$$
\tilde{V}_{k, p}\triangleq\lambda I+\sum_{t=p}^{k} \tilde{Z}_{t, p} \tilde{Z}_{t, p}^{\tr}, \quad \bar{V}_{k, p}\triangleq\lambda D_{p}^{-2}+\sum_{t=p}^{k} Z_{t, p} Z_{t, p}^{\tr}.
$$
It is clear that both are positive definite. 
The modified cross-term is defined as
$
\tilde{S}_{k, p}\triangleq\sum_{l=p}^{k} y_{l} \tilde{Z}_{l, p}^{\tr}=S_{k, p} D_{p}.$ 
Then, the update of the regression coefficient in \eqref{eq:regression-update} becomes
$
\tilde{G}_{k, p}=\tilde{S}_{k, p} \tilde{V}_{k, p}^{-1}
$
and the prediction of \texttt{OPF} at time step $k+1$ is given in \eqref{eq:OPF-prediction}. 

We now have the following equivalence 
    \begin{align}
    \tilde{S}_{k,p} \tilde{V}_{k,p}^{-1}\tilde{Z}_{k+1,p}
    =&
S_{k, p}\left(\lambda D_{p}^{-2}+\sum_{k=1}^{k} Z_{k,p} Z_{k, p}^{\tr}\right)^{-1} Z_{k+1, p} \nonumber \\ 
=& S_{k, p} \bar{V}_{k, p}^{-1} Z_{k+1, p}. \label{equiv}
\end{align}
Thus, applying our exponentially forgetting factor $\gamma$ is mathematically equivalent to replacing the regularization term $\lambda I$ with $\lambda D_p^{-2}$. 
{More specifically, $\tilde{G}_{k,p}$ in \eqref{eq:regression-update} is the solution to the following generalized ridge regression 
\begin{equation}\label{ridgeRegre}
    \min_{G \in \mathbb{R}^{m \times mp}} \,\,  \sum_{t=p}^k \left\|y_{t}-G{Z}_{t,p}\right\|_F^2+\left\|\lambda GD_p^{-1}\right\|_F^2. 
\end{equation}
The original ridge regression for \eqref{regression} is similar to \eqref{ridgeRegre} but with a usual regularization $\left\|\lambda G\right\|_F^2$. 
For \eqref{ridgeRegre}, we can see that the forgetting factor $\gamma$ acts as a regularization $\left\|\lambda GD_p^{-1}\right\|_F^2$. Indeed, this regularization injects our prior belief that the ground truth of $G$ has an exponentially decaying structure as in \eqref{eq:regressor}, with the penalty term $\frac{1}{\gamma^{t}}$ effectively reducing the effect of overfitting.
}

\section{Logarithmic Regret of the \texttt{OPF} Algorithm}

\label{Section:Regret-guarantee}

We here prove that with high probability, the~regret of our \texttt{OPF} in \Cref{algPrediction} is logarithmic of the order $\log^{3} N$, which greatly improves the existing results (see \Cref{sample-table}).

\subsection{A Sharper Logarithmic Regret}
Recall that the regret $\mathcal{R}_N$ in \eqref{regret} measures against the optimal Kalman filter's prediction with true model knowledge. We have the following logarithmic guarantee for our proposed \texttt{OPF} in \Cref{algPrediction}. 
\begin{theorem}\label{thm1}
    Consider the linear stochastic system \eqref{linearSystem} with $Q \succ 0, R \succ 0$, and suppose \Cref{aspOb-main-text,aspDiagonal} hold. Choose the forgetting factor $\gamma \in (\rho(A-LC),1]$, and the parameters in \Cref{algPrediction}  as  $$\beta=\frac{\Omega(\kappa)}{\log (1 / \rho(A-L C))},\quad 
    T_{\mathrm{init}} = \operatorname{poly}\left(\beta,\log\left(\frac{1}{\delta}\right)\right), $$ where $\kappa$ represents the order of the largest Jordan block of $A$ corresponding to the eigenvalue 
$1$. Fix a horizon $N > T_{\mathrm{init}}$ and a failure probability $\delta > 0$. Then  with probability at least $1-\delta$, we have 
    $$
    \mathcal{R}_{N} \leq \operatorname{poly}\left(M,\beta,\log(1/\delta)\right) \mathcal{O}\left(\log^{3} N\right),$$
    where $M$ is a constant only related to the system parameter.
\end{theorem}

    Achieving a logarithmic regret is not surprising since we learn the regression model \eqref{regression-balanced}, which is in the framework of online least-squares \cite{draper1998applied}.     
    Given the fact that $(A-LC)^p$ is exponentially decaying and the innovation noise $e_k$ is mutually uncorrelated Gaussian, selecting a large $p=O(\log T_l)$ in each epoch is sufficient to obtain an online predictor $\tilde{y}_k$ with logarithmic regret. In \cite{hazan2017learning, li2024regret}, the system noise is not guaranteed to be i.i.d. Gaussian, hence the algorithms therein do not have logarithmic regrets. In two recent studies \cite{tsiamis9894660,rashidinejad2020slip} which focused on the original unbalanced regression model \eqref{regression}, the regret bound is shown to be $O\left(\log^{6} N\right)$ and $O\left(\log^{11} N\right)$, respectively. 
    In \Cref{thm1}, we have established a much shaper bound compared with the existing results; also see \Cref{sample-table} for a comparison. Our proof techniques involve new tools from Hanson-Wright inequality-based martingale analysis and self-cancellation property of accumulation error.
    
    Our \Cref{thm1} confirms that our idea of forgetting factor effectively balances the regression model with no information loss. Note that the forgetting factor $\gamma$ may not directly reduce the order of the regret to $\log(N)$, but it can effectively reduce the {accumulation} error, and balance a better trade-off between {regression} error and {regularization} error, thus significantly reducing some constant in the logarithmic regret. In \Cref{secDiscussion}, we provide further discussions on the effect of the forgetting factor. 

\subsection{Proof Sketches for the Shaper Regret} \label{subsection:proof-sketch}

In this subsection, we provide proof sketches for \Cref{thm1}. Rather than presenting all proof details (which are given in \Cref{secMainProof}), we here highlight the main theoretical tools that achieve a sharper regret bound compared with the state-of-the-art in \Cref{sample-table}.

\textbf{Step 1: Accumulation error and martingale terms.} From general online linear regression techniques \cite{tsiamis9894660, rashidinejad2020slip}, we first divide the regret $\mathcal{R}_N$ into two parts, i.e.,\vspace{-5pt}
\begin{align} \label{eq:decomposition-regret}
    \mathcal{R}_{N} \triangleq& \sum_{k=T_{\text {init }}}^{N}\left\|y_{k}-\tilde{y}_{k}\right\|_2^{2}\;-\sum_{k=T_{\text {init }}}^{N}\left\|y_{k}-\hat{y}_{k}\right\|_2^{2} \nonumber \\
    =&\underbrace{\sum_{k=T_{\text {init }}}^{N}\left\|\hat{y}_{k}-\tilde{y}_{k}\right\|_{2}^{2}}_{\mathcal{L}_{N}}\;+\;2 \underbrace{\sum_{k=T_{\text {init }}}^{N} e_{k}^{\tr}\left(\hat{y}_{k}-\tilde{y}_{k}\right)}_{\text {martingale term }} .
\end{align}
The first part is the accumulation of the gap between $\hat{y}_{k}-\tilde{y}_{k}$, while the second part is a cross term between the innovation $e_k$ and the gap $\hat{y}_{k}-\tilde{y}_{k}$. Since $e_k$ is i.i.d. Gaussian, from the self-normalized martingale theory \cite{abbasi2011improved, tsiamis2019finite}, it is standard to bound
\begin{equation} \label{eq:bound-marignale-term}
\sum_{k=T_{\text {init }}}^{N} e_{k}^{\tr}\left(\hat{y}_{k}-\tilde{y}_{k}\right)=\tilde{O}\left(\sqrt{\mathcal{L}_N}\right)=o\left(\mathcal{L}_N\right),
\end{equation}
i.e., the cross term is dominated by the accumulation term $\mathcal{L}_N$. 

In the rest of the proof, we will tightly bound $\mathcal{L}_N$.

\vspace{3pt}

\textbf{Step 2: Regularization, regression and bias errors.} Following standard  linear regression techniques \cite{hoerl1970ridge, abbasi2011improved, hazan2017learning}, we can divide the gap $\hat{y}_{k}-\tilde{y}_{k}$ into three parts: 
\begin{itemize}
    \item  1) the \textit{regularization} error induced by $\lambda D_p^{-2}$, 
\item 2) the \textit{regression} error induced by Gaussian error $e_k$, and 
\item 3) the \textit{bias} error induced by the long-term memory $b_{k,p}=C(A-K C)^{p} \hat{x}_{k-p}$. 
\end{itemize}
Consequently, we rewrite the gap $\tilde{y}_{k}-\hat{y}_k$ at each time as 
\begin{equation}\label{eq:gap-yk-decomposition}
    \tilde{y}_{k+1}-\hat{y}_{k+1}=  \underbrace{\sum_{l=p}^{k} b_{l, p} Z_{l, p}^{\tr} \bar{V}_{k, p}^{-1} Z_{k+1, p}-b_{k+1, p}}_{\text{Bias error}} +\underbrace{\sum_{l=p}^{k} e_{l} Z_{l, p}^{\tr} \bar{V}_{k, p}^{-1} Z_{k+1, p}}_{\text{Regression error}}-\underbrace{\lambda G_{p} D_{p}^{-2} \bar{V}_{k, p}^{-1} Z_{k+1, p}}_{\text{Regularization error}}. 
\end{equation}
One main difference between  \eqref{eq:gap-yk-decomposition} with that in \cite{tsiamis9894660, rashidinejad2020slip} is the scaling matrix $D_p$ induced by the forgetting factor $\gamma$ (see \eqref{equationDp}) will simultaneously affect the regularization error and the inverse of the Gram matrix $\bar{V}_{k,p}^{-1}$. We will provide a detailed discussion on the effect of $D_p$ for a better trade-off between regularization and accumulation in \Cref{secDiscussion}.

From \cref{eq:gap-yk-decomposition}, we can bound $\mathcal{L}_N$ by
    \begin{align} \label{wholeRegret}
\mathcal{L}_{N}=&\sum_{k=T_{\text{init}}}^{N}\left\|\tilde{y}_{k+1}-\hat{y}_{k+1}\right\|_{2}^{2} \nonumber\\
\leq&6\sup_{T_{\text{init}}\leq k\leq N
}\left(\left\| \lambda G_{p} D_{p}^{-2} \bar{V}_{k, p}^{-\frac{1}{2}}\right\|_{2}^{2}+\left\| B_{k, p} \bar{Z}_{k, p}^{\tr} \bar{V}_{k, p}^{-\frac{1}{2}} \right\|_{2}^{2}+\left\|E_{k, p} \bar{Z}_{k, p}^{\tr} \bar{V}_{k, p}^{-\frac{1}{2}}\right\|_{2}^{2}\right)
\\&\times\sum_{k=T_{\text{init}}}^{N}\left\|\bar{V}_{k, p}^{-\frac{1}{2}} Z_{k+1,p}\right\|_{2}^{2} \nonumber+2\sum_{k=T_{\text{init}}}^{N}\left\|b_{k+1, p}\right\|_{2}^{2}, 
\end{align}
where $B_{k,p}\triangleq\begin{bmatrix}
    b_{p,p},\dots,b_{k,p}
\end{bmatrix}$, $E_{k,p}\triangleq\begin{bmatrix}
    e_p,\dots,e_k
\end{bmatrix}$ and $\bar{Z}_{k,p}\triangleq\begin{bmatrix}
    Z_{p,p},\dots,Z_{k,p}
\end{bmatrix}$ are the collections of all past bias $b_{l,p}$, errors $e_l$ and samples $Z_{l,p}$, respectively. 

The bias factor $\big\| B_{k, p} \bar{Z}_{k, p}^{\tr} \bar{V}_{k, p}^{-\frac{1}{2}} \big\|_{2}^{2}$ can be bounded by choosing a sufficiently large $\beta$, as presented in \Cref{thm1}.  

For the regularization factor $\big\| \lambda G_{p} D_{p}^{-2} \bar{V}_{k, p}^{-\frac{1}{2}}\big\|_{2}^{2}$, observing that
$$
\lambda G_{p} D_{p}^{-2} \bar{V}_{k, p}^{-1} D_{p}^{-2} G_p^\tr\leq G_pD_p^{-1}D_p^{-1}G_p^\tr,
$$ 
we~have
\begin{equation}
\left\| \lambda G_{p} D_{p}^{-2} \bar{V}_{k, p}^{-\frac{1}{2}}\right\|_{2}^{2}\leq\sum_{i=1}^{p}M\left(\frac{\rho(A-LC)}{\gamma}\right)^{i-1},
\end{equation}
where $M$ is a constant only depending on system matrices. 

For the regression factor $\big\|E_{k, p} \bar{Z}_{k, p}^{\tr} \bar{V}_{k, p}^{-\frac{1}{2}}\big\|_{2}^{2}$, note that the diagonal elements of $E_{k, p} \bar{Z}_{k, p}^{\tr} \bar{V}_{k, p}^{-1}\bar{Z}_{k, p}E_{k, p}^\tr$ are quadratic forms of Gaussian random variables, hence we can adapt the \textit{Hanson-Wright inequality} \cite{hanson1971bound, vershynin2018high} to provide a bound with order of $\text{log}(N)$: 
\begin{equation} \label{eq:bound-Hanson-Wright}
\sup_{T_{\text{init}}\leq k \leq N}\left\|E_{k, p} \bar{Z}_{k, p}^{\tr} \bar{V}_{k, p}^{-\frac{1}{2}}\right\|_{2}^{2} \!\leq\! \operatorname{poly}\left(\!\log\frac{1}{\delta}\!\right)\log (N).
\end{equation}
In \cite{tsiamis9894660,rashidinejad2020slip}, self-normalized martingale theory \cite{abbasi2011improved} is used to bound $\big\|E_{k, p} \bar{Z}_{k, p}^{\tr} \bar{V}_{k, p}^{-\frac{1}{2}}\big\|_{2}^{2}$ in terms of the determinant of $\bar{V}_{k, p}$, i.e,
$$
\sup_{T_{\text{init}}\leq k \leq N}\left\|E_{k, p} \bar{Z}_{k, p}^{\tr} \bar{V}_{k, p}^{-\frac{1}{2}}\right\|_{2}^{2}\leq \log\frac{\det(\bar{V}_{k,p})}{\delta\det(\lambda D_p^{-2})},
$$
with high probability $1-\delta$. However, the combination of persistent excitation of  $\bar{V}_{k,p}\ge{\frac{\sigma_R}{4}kI}$ and high dimension $p=O(\log(N))$ lead to a conservative bound of $\log\det(\bar{V}_{k,p})$ with order  $\log^2(N)$, which is not sharp.  
Finally, we analyze the bound for the accumulation factor
$\sum_{k=T_{\text{init}}}^{N}\big\|\bar{V}_{k, p}^{-\frac{1}{2}} Z_{k+1,p}\big\|_{2}^{2}$. Since the term $p$ varies with different epochs, we denote $N_E$ as the number of all epochs and let $T_l=2^{l-1}T_{\text{init}}+1$ be the initial time step at the $l$-th epoch and $p_l=\beta\log T_l$ as the past horizon length in the $l$-th epoch. We have $N=2^l T_{\text{init}}$ and
$$
\begin{aligned}
    \sum_{k=T_{\text{init}}
    }^{N}\left\|\bar{V}_{k, p}^{-\frac{1}{2}} Z_{k+1,p}\right\|_{2}^{2}\leq \sup_{T_{\text{init}}\leq k\leq N}\left\|\bar{V}_{k, p}^{-\frac{1}{2}} \bar{V}_{k+1, p}^{\frac{1}{2}} \right\|_2^2\times\sum_{l=1}^{N_E}\sum_{k=T_l}^{2T_l-2}\left\|\bar{V}_{k+1, p}^{-\frac{1}{2}} Z_{k+1,p}\right\|_{2}^{2}.
\end{aligned}
$$
For the cross term $\big\|\bar{V}_{k, p}^{-\frac{1}{2}} \bar{V}_{k+1, p}^{\frac{1}{2}} \big\|_2^2$, we can utilize the persistent-excitation condition 
to provide a uniform constant bound $M_1$ to all cross terms, i.e., 
$$
\big\|\bar{V}_{k, p}^{-\frac{1}{2}} \bar{V}_{k+1, p}^{\frac{1}{2}} \big\|_2^2\leq M_1
$$
for all $k\ge T_{\text{init}}$ with high probability. For the accumulation factor $\sum_{k=T_l}^{2T_l-2}\big\|\bar{V}_{k+1, p}^{-\frac{1}{2}} Z_{k+1,p}\big\|_{2}^{2}$,
we can provide an upper bound for the accumulation term in one epoch with the determinant of the Gram matrix, i.e.,
$$
\begin{aligned}
    \sum_{k=T_l}^{2T_l-2}\left\|\bar{V}_{k+1, p}^{-\frac{1}{2}} Z_{k+1,p}\right\|_{2}^{2}\leq& \log\frac{\text{det}\tilde{V}_{2T_l-1,p}}{\text{det}\tilde{V}_{T_l,p}}.
\end{aligned}
$$
The sum of the determinant term is bounded by $\log\det\tilde{V}_{N,p}$ and the additional tail is bounded by $O(\log^2 N)$, i.e.,
\begin{equation} \label{eq:bound-common-multiplier}
\sum_{l=1}^{N_E}\log\frac{\text{det}\tilde{V}_{2T_l-1,p}}{\text{det}\tilde{V}_{T_l,p}}\leq \log\det\tilde{V}_{N,p}+O(\log^2N).
\end{equation}

One technical point is that the determinant $\det\tilde{V}_{T_l,p_l}$ and $\det\tilde{V}_{2T_{l-1}-1,p_{l-1}}$ will cancel with each other with small residue (see \Cref{secMainProof} for more details).  
We further~prove that with  marginally stable $A$, i.e. $\rho(A)=1$, the accumulation  term $\sum_{k=T_{\text{init}}}^{N}\big\|\bar{V}_{k, p}^{-\frac{1}{2}} Z_{k+1,p}\big\|_{2}^{2}$ is still with the order of $\log^2(N)$, rather than $\log^4 (N)$ in \cite{tsiamis9894660}.

\textbf{Step 3: Add all error bounds.} Putting the bounds~in~\eqref{eq:bound-Hanson-Wright} and \eqref{eq:bound-common-multiplier} into \eqref{wholeRegret}, we derive a bound for $\mathcal{L}_N$ of order $\log^3 N$. Finally, as $\mathcal{L}_N$ dominates the martingale~term~in \eqref{eq:bound-marignale-term}, we derive the regret bound for  $\mathcal{R}_N$ of order~$\log^3 N$ from \eqref{eq:decomposition-regret}. This completes the proof sketch of \cref{thm1}. 
    
\begin{remark}
Our proof shows that the ``doubling trick'' that handles long-term memory will not~induce additional regret. The cancellation effect between the successive epochs guarantees good regret performance.
With new techniques, we provide a sharper bound on the regret $\mathcal{L}_N$ from $O(\log^6 N)$ \cite{tsiamis9894660} to $O(\log^3 N)$. 
When implementing \Cref{algPrediction}, the parameter $p$ should be an integer. Thus, we need to use  $p=\lceil\beta\log T_l\rceil$, but rounding up to an integer will not affect the order of regret. \hfill $\square$ 
\end{remark}

%% file: Section-V-v0.tex
\section{Numerical Experiments}\label{secSimulation}

In this section, 
we compare our proposed \texttt{OPF} in \Cref{algPrediction} with the algorithm in \cite{tsiamis9894660} and traditional forgetting techniques \cite{jacobsen2024online}. Further experiments are given in \Cref{secAdditionalExperiments}.  

\begin{figure}[t]
    \centering
    \setlength{\abovecaptionskip}{0pt}
    \subfigure[Our Forget]{\includegraphics[width=0.5\textwidth]{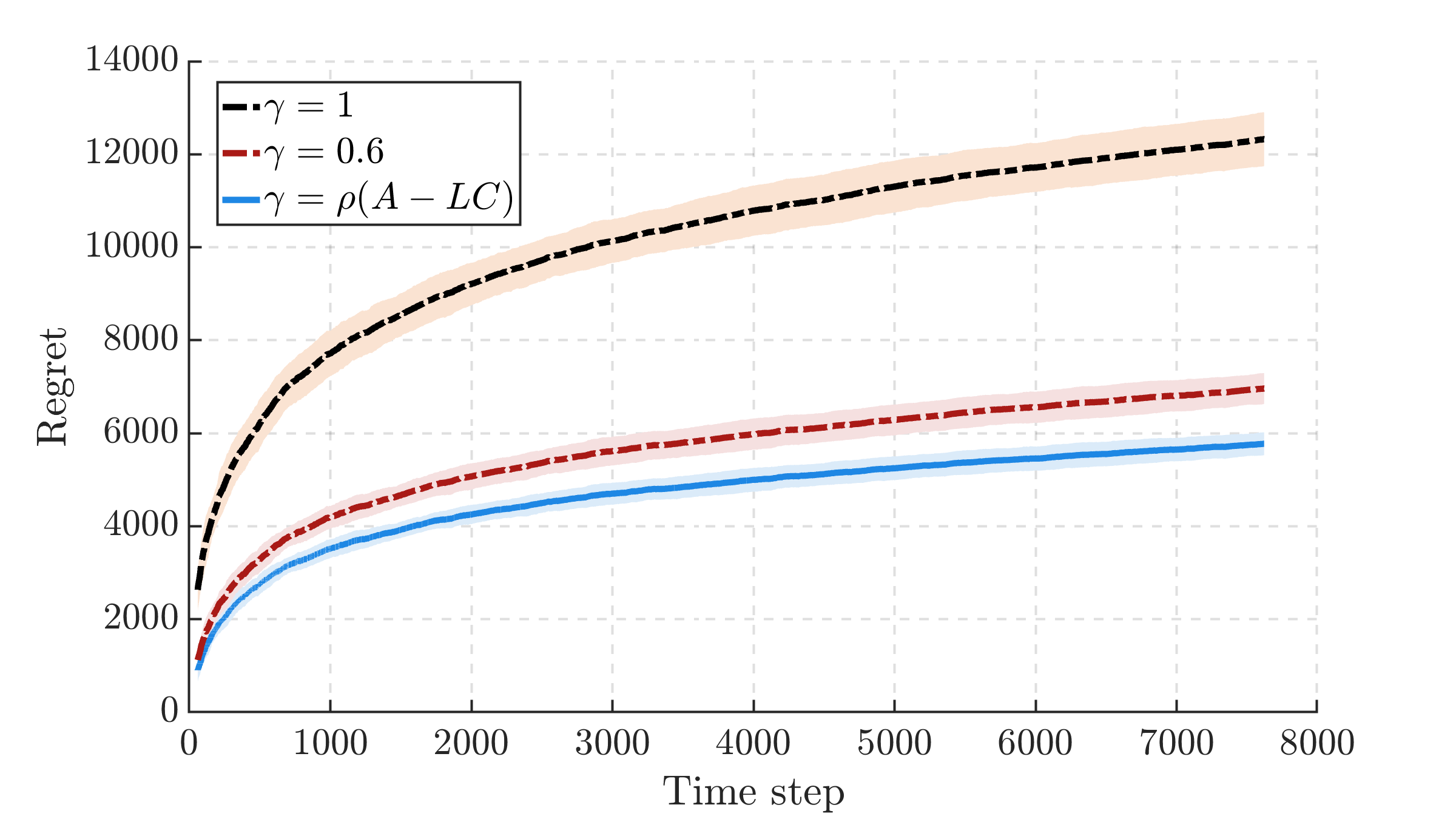}
    \label{figComparison}}\subfigure[Traditional Forget]{\includegraphics[width=0.5\textwidth]{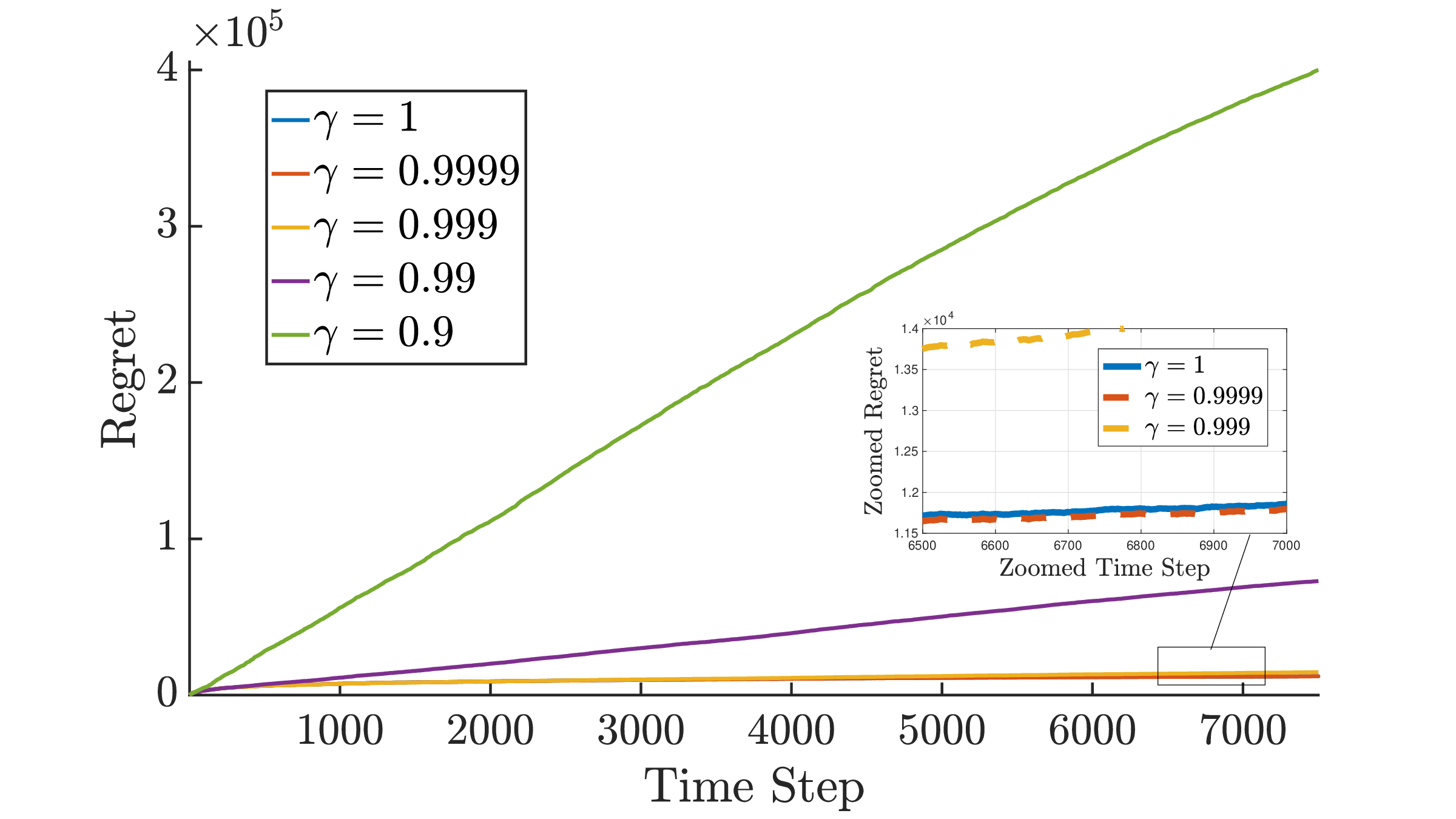}
\label{figDiscussionForget}}
    \caption{Performance comparison of online prediction with different forgetting strategies: (a) our proposed forgetting; (b) traditional forgetting in \cite{jacobsen2024online}.}

    \vspace{0mm}
\end{figure}

We first consider an augmented form of the dynamical system model from \cite[Section V]{tsiamis9894660}, where the system parameters are defined as
$$
A=I_3\otimes \Tilde{A}, C=I_3\otimes
\Tilde{C}, Q = \begin{bmatrix}
    I_3&0.2I_3&0.2I_3\\
    0.2I_3&I_3&0.2I_3\\
    0.2I_3&0.2I_3&I_3
\end{bmatrix}$$
where
$$
\Tilde{A}=\begin{bmatrix}
1 & 1 & 0 \\
0 & 1 & 1 \\
0 & 0 & 0.9
\end{bmatrix},\;\; \Tilde{C}=\begin{bmatrix}
    1&0&0
\end{bmatrix},\;\;R=I_3.
$$ 
This can be interpreted as a 3-D target tracking problem with an unknown correlated dynamic model \cite{cattivelli2010diffusion}.
The parameters in our \texttt{OPF} algorithm are chosen as 
$$\beta=2.5,T_{\text{init}}=60, \lambda=1, N_E = 7.$$ 
Accordingly, the total horizon is $7680$. The comparison result with different forgetting factor $\gamma$ for \Cref{algPrediction} is shown in \Cref{figComparison}, where $\gamma=1$ represents the performance of the algorithm in \cite{tsiamis9894660}. 
From  \Cref{figComparison}, we can see that the performance of $\gamma=\rho(A-LC)$ performs better than other forgetting factor $\gamma\in(\rho(A-LC),1]$. 

We also implemented the traditional forgetting techniques \eqref{traditionalForget} from  \cite{jacobsen2024online} for online prediction of marginally stable systems. As we can see from \Cref{figDiscussionForget}, only for $\alpha=0.9999$, almost close to $1$, the performance of the online prediction algorithm is slightly improved. When $\alpha$ is chosen to be 0.99, the forgetting-based method performs poorly compared with the non-forgetting case $\alpha=1$, which verifies our discussions in \Cref{rmkTraditional}. Hence, the forgetting technique proposed in this paper is a more effective way to improve the performance of online prediction. This is incorporated in our \texttt{OPF} algorithm.

\begin{figure}[t]
    \centering
    \subfigure[Accumulation Error]{\includegraphics[width=0.45\textwidth]{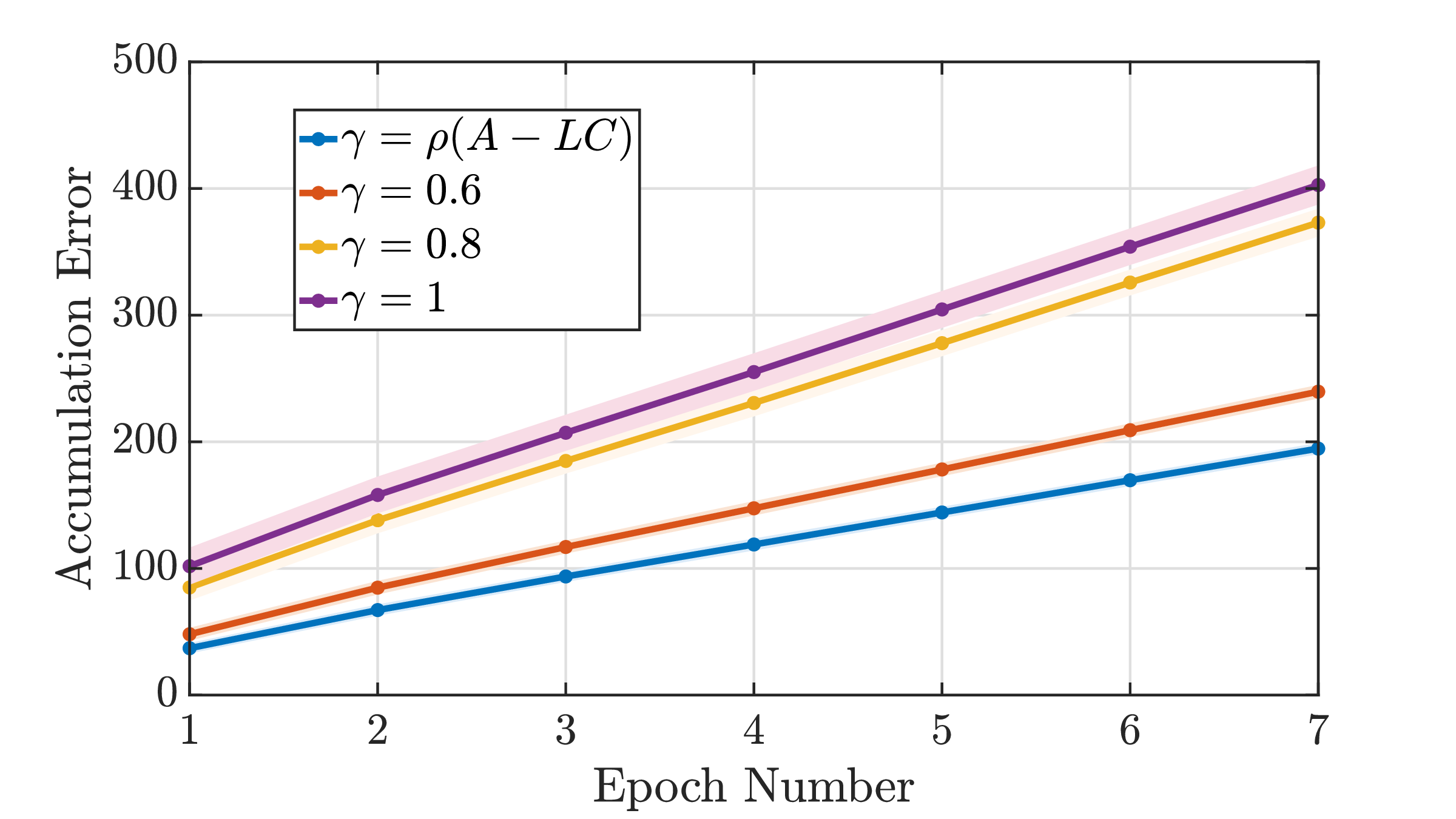}
    \label{figAccumulation}
    }
    \subfigure[Trade-Off]{\includegraphics[width=0.45\textwidth]{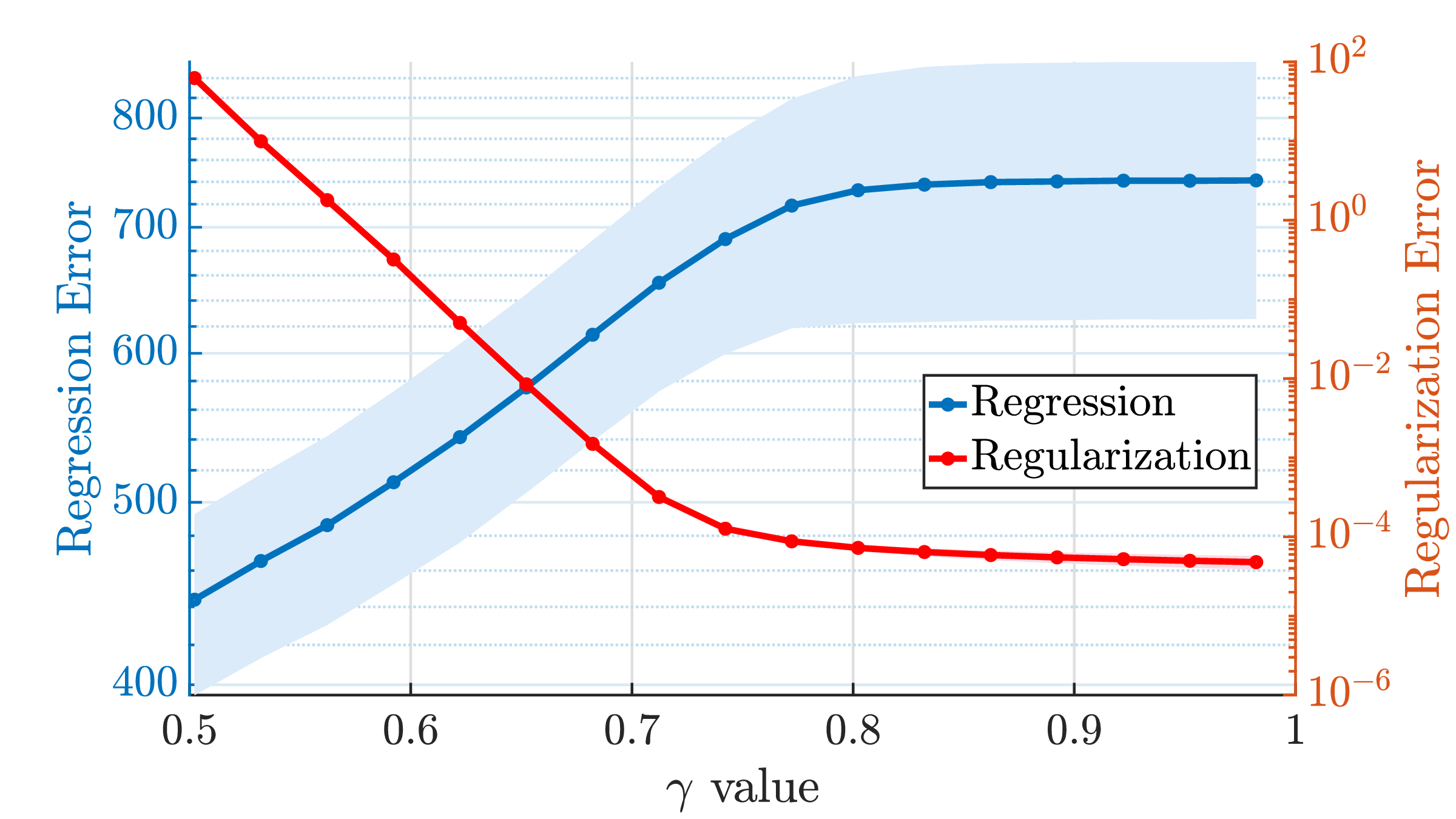}
    \label{figTradeoff}}
    \caption{Effect of our forgetting factor on different parts of regret $\mathcal{R}_N$ in \eqref{wholeRegret}.
    }
    \label{figVerification}

    \vspace{0mm}
\end{figure}

In our second numerical experiment, we verify that the forgetting factor $\gamma$ can effectively reduce the regret $\mathcal{R}_N$. In \cref{secDiscussion}, we prove that the forgetting factor can reduce the accumulation error $\sum_{k=T_{\textbf{init}}}^{N}\big\|\bar{V}_{k, p}^{-\frac{1}{2}} Z_{k+1,p}\big\|$ and regression error $\big\|E_{k, p} \bar{Z}_{k, p}^{\tr} \bar{V}_{k, p}^{-\frac{1}{2}}\big\|_{2}^{2}$, while not increasing the regularization error
$\big\| \lambda G_{p} D_{p}^{-2} \bar{V}_{k, p}^{-\frac{1}{2}}\big\|_{2}^{2}$ too much. In \cref{figAccumulation}, we choose the accumulation error at the end of each epoch for different $\gamma$ as the performance index. In \cref{figTradeoff}, we choose the regression error and regularization error for different $\gamma$ at time step $N$ for comparison, where the shaded area is calculated with one standard deviation. The result in \cref{figVerification} is consistent with our claim in \cref{secDiscussion}.

%% file: Section-IV-v1.tex
\section{Regret Improvement with Forgetting}
\label{secDiscussion}

In this section, we will analyze the effect of forgetting factor $\gamma$ on the regret $\mathcal{R}_N$. 
From the proof of Theorem \ref{thm1}, we can have an  intuitive idea that the forgetting factor $\gamma$ influences the regret $\mathcal{R}_N$ by affecting  
\begin{itemize}
\item the regularization term $\big\| \lambda G_{p} D_{p}^{-2} \bar{V}_{k, p}^{-\frac{1}{2}}\big\|_{2}^{2}$, 
    \item regression term $\big\|E_{k, p} \bar{Z}_{k, p}^{\tr} \bar{V}_{k, p}^{-\frac{1}{2}}\big\|_{2}^{2}$ and 
    \item accumulation term $\sum_{k=T_{\text{init}}}^{N}\big\|\bar{V}_{k, p}^{-\frac{1}{2}} Z_{k+1,p}\big\|_{2}^{2}$. 
\end{itemize} 
Moreover, the theoretical analysis shows that the term $\big\|E_{k, p} \bar{Z}_{k, p}^{\tr} \bar{V}_{k, p}^{-\frac{1}{2}}\big\|_{2}^{2}$ and $\sum_{k=T_{\text{init}}}^{N}\big\|\bar{V}_{k, p}^{-\frac{1}{2}} Z_{k+1,p}\big\|_{2}^{2}$ are dominated by $\bar{\sigma}_{\bar{R}}\text{tr}\big(\bar{Z}_{k, p}^{\tr} \bar{V}_{k, p}^{-{1}}\bar{Z}_{k, p}\big)$ and $\det(\tilde{V}_{N,p})$ respectively. 

Therefore, in this subsection, to fully illustrate the essence of the effect of the forgetting factor, our analysis will mainly be based on 
the relationship between $\gamma$ and the dominant terms.

\subsection{Effect of forgetting factor $\gamma$}\label{subsecTradeoff}
For the regularization term, we have
$$
\big\| \lambda G_{p} D_{p}^{-2} \bar{V}_{k, p}^{-\frac{1}{2}}\big\|_{2}^{2}\leq\sum_{i=1}^{p}M\left(\frac{\rho(A-LC)}{\gamma}\right)^{i-1}.
$$
Note that for the special case $\gamma=\rho(A-LC)$, we have
$$
\big\| \lambda G_{p} D_{p}^{-2} \bar{V}_{k, p}^{-\frac{1}{2}}\big\|_{2}^{2}\leq Mp\leq M\beta \log(N).
$$
while for the general case when $\gamma>\rho(A-LC)$, the regularization term is uniformly bounded by $\frac{M\gamma}{\gamma-\rho(A-LC)}$ for all $N$. Hence for all $\gamma\ge \rho(A-LC)$, the effect of $\gamma$ to increase the regularization error will be bounded by $O(\log N)$, and it will only have a sharp increase when $\gamma$ is close to $\rho(A-LC)$.

While for the term $\big\|E_{k, p} \bar{Z}_{k, p}^{\tr} \bar{V}_{k, p}^{-\frac{1}{2}}\big\|_{2}^{2}$, which is dominated by $\bar{\sigma}_{\bar{R}}\text{tr}\big(\bar{Z}_{k, p}^{\tr} \bar{V}_{k, p}^{-{1}}\bar{Z}_{k, p}\big)$. We have
$$
\text{tr}\big(\bar{Z}_{k, p}^{\tr} \bar{V}_{k, p}^{-{1}}\bar{Z}_{k, p}\big)=\text{tr}\big(\bar{V}_{k, p}^{-{1}}\bar{Z}_{k, p}\bar{Z}_{k, p}^{\tr}\big)=\text{tr}\big(I_{mp}-\lambda \tilde{V}_{k,p}^{-1}\big).
$$
For the term $\text{tr}\big(\lambda \tilde{V}_{k,p}^{-1}\big)$, we have
$$
\lambda I+D_p\bar{Z}_{k,p}\bar{Z}_{k,p}^\top D_p\leq \lambda I+\operatorname{poly}\left(M,\beta,\log\frac{1}{\delta}\right)k^{2\kappa} D_p^2.$$ with high probability $1-\delta$, where $M$ is only related to system parameter. Then we have
$$
\text{tr}\big(\bar{Z}_{k, p}^{\tr} \bar{V}_{k, p}^{-{1}}\bar{Z}_{k, p}\big)\!\leq \!\sum_{l=0}^{p-1}m-\frac{m\lambda}{\lambda \!+ \!\operatorname{poly}\!\left(M,\beta,\log\frac{1}{\delta}\right)\!k^{2\kappa}\gamma^{2l}}.$$
The above inequality shows that the forgetting factor $\gamma$ can reduce the regression term $\big\|E_{k, p} \bar{Z}_{k, p}^{\tr} \bar{V}_{k, p}^{-\frac{1}{2}}\big\|_{2}^{2}$through reducing $\operatorname{poly}\left(M,\beta,\log\frac{1}{\delta}\right)k^{2\kappa}\gamma^{2l}$. 
Consider the case $\gamma   =\rho(A-LC)$, from the selection of parameter $\beta$ and $p\leq \beta \log k$, we have
$$
\rho(A-LC)^p = O(\frac{1}{k^{2\kappa+1}}).
$$
Hence for any $l\ge \frac{p}{2}$, there is 
$$
\frac{m\lambda}{\lambda + \operatorname{poly}\left(M,\beta,\log\frac{1}{\delta}\right)k^{2\kappa}\gamma^{2l}}\leq \frac{m\lambda}{\lambda+O(1/k)}\approx m
$$
and
$$
\text{tr}\big(\bar{Z}_{k, p}^{\tr} \bar{V}_{k, p}^{-{1}}\bar{Z}_{k, p}\big)\lesssim \frac{mp}{2}.
$$
Hence we can find that the reduction effect of the forgetting factor to the regression term is efficient. Hence we have the intuitive idea that
the forgetting factor can balance a better trade-off between the regularization error and the regression error, which will also be verified with numerical simulation in \Cref{secSimulation}.

We then consider the connection between $\gamma$ and $\det(\tilde{V}_{N,p})$. With similar technique, for the case $\gamma=\rho(A-LC)$, we can obtain
$$
\begin{aligned}
    \det(\tilde{V}_{N,p})\leq& \prod_{l=0}^{p-1}\left(\lambda + \operatorname{poly}\left(M,\beta,\log\frac{1}{\delta}\right)N^{2\kappa}\gamma^{2l}\right)\\
    \lesssim& \lambda^{p/2}\prod_{l=0}^{p/2}\left(\lambda + \operatorname{poly}\left(M,\beta,\log\frac{1}{\delta}\right)N^{2\kappa}\gamma^{2l}\right).
\end{aligned}
$$
Hence we can find that the effect of $\gamma=\rho(A-LC)$ to reduce the regression term and accumulation term is at least proportional to the case with $\gamma=1$, which verifies the effectiveness of applying forgetting factor from a theoretical perspective.

\subsection{Robust analysis for the selection of parameter $\gamma$} \label{sectionRobust}
Note that when $\gamma>\rho(A-LC)$, the logarithm regret $\mathcal{R}_N$ of Algorithm \ref{algPrediction} with optimal Kalman filter can be guaranteed. In this subsection, we aim to provide essential discussion on the robustness of choosing parameter $\gamma$, which indicates there are still some margin $\epsilon$ to allow $\gamma\leq \rho(A-LC)-\epsilon$ and still guarantee the regret $\mathcal{R}_N$ to be sub-linear, i.e., $\mathcal{R}_N=\text{poly}\left(\frac{1}{\delta}\right)o(N),$ for all time step $N$. 

We first consider the regularization error $\big\| \lambda G_{p} D_{p}^{-2} \bar{V}_{k, p}^{-\frac{1}{2}}\big\|_{2}^{2}$,
note that with the previous analysis, there is
$$\left\| \lambda G_{p} D_{p}^{-2} \bar{V}_{k, p}^{-\frac{1}{2}}\right\|_{2}^{2}\leq\sum_{i=1}^{p}M\left(\frac{\rho(A-LC)}{\gamma}\right)^{i-1},$$
where $M$ is only related to system parameters. Then consider $\gamma = \frac{1}{1+c}\rho(A-LC),\; c>0$. Note that when $c< e^{\frac{1}{\beta^\prime}}-1$ we have
$$
\left\| \lambda G_{p} D_{p}^{-2} \bar{V}_{k, p}^{-\frac{1}{2}}\right\|_{2}^{2}\leq M\sum_{i=1}^p (1+c)^i\leq \frac{e^{\frac{p}{\beta^\prime}}}{e^{\frac{1}{\beta^\prime}}-1}.
$$
Together with $p\leq \beta \log k$, we can obtain
$$
\sup_{T_{\text{init}}\leq k\leq N
}\left\| \lambda G_{p} D_{p}^{-2} \bar{V}_{k, p}^{-\frac{1}{2}}\right\|_{2}^{2}\leq O(N^{\frac{\beta}{\beta^\prime}}).$$
With a similar proof as Theorem \ref{thm1}, we can prove that all the other terms in the regret upper bound \eqref{wholeRegret} are with the same order as $\operatorname{poly}(\log N)$, hence we can obtain that for all $\beta^\prime > \beta, c\leq e^{\frac{1}{\beta^\prime}}-1$ and $\gamma=\frac{1}{1+c}\rho(A-LC)$, we have
$$
\mathcal{R}_N=\operatorname{poly}(M,\log(1/\delta))O(N^{\frac{\beta}{\beta^\prime}}\operatorname{poly}(\log N))=o(N).$$ Hence the robust margin for the selection of $\gamma$ is not larger than
$$\epsilon\triangleq \rho(A-LC)-\gamma=\left(1-\frac{1}{e^{\frac{1}{\beta}}}\right) \rho(A-LC). $$